\documentclass[10pt,journal,compsoc]{IEEEtran}
\usepackage{amsmath,amsfonts,amssymb,bbm}
\usepackage{algorithmic}
\usepackage{algorithm}
\usepackage{array}
\usepackage{textcomp}
\usepackage{stfloats}
\usepackage{url}
\usepackage{verbatim}
\usepackage{graphicx}
\usepackage{cite}
\usepackage{enumitem}
\usepackage{booktabs}
\usepackage{multicol}
\usepackage{multirow}
\usepackage{xcolor} 
\usepackage{cases}
\usepackage{caption}
\usepackage{subcaption}
\usepackage[pagebackref,breaklinks,colorlinks]{hyperref}
\hypersetup{linkcolor=red,anchorcolor=blue,citecolor=cyan,filecolor=blue,urlcolor=blue}

\usepackage[capitalize]{cleveref}
\crefname{section}{Sec.}{Secs.}
\Crefname{section}{Section}{Sections}
\Crefname{table}{Table}{Tables}
\crefname{table}{Tab.}{Tabs.}

\usepackage{orcidlink}

\newcommand{\relleft}{\texttt{left}}

\newcommand{\sr}{SR\textsubscript{2D}}
\newcommand{\etal}{\textit{et al.\ }}
\newcommand{\eg}{\textit{e.g.\ }}
\newtheorem{definition}{Definition}
\definecolor{darkblue}{rgb}{0.2, 0, 0.8}
\definecolor{darkgreen}{rgb}{0.25, 0.75, 0.3}
\definecolor{darkpurple}{rgb}{0.6, 0, 0.75}

\begin{document}

\title{Benchmarking Spatial Reasoning Abilities of Text-to-Image Generative Models}

\author{
    Tejas Gokhale, Hamid Palangi, Besmira Nushi, Vibhav Vineet, 
    \\Eric Horvitz, Ece Kamar, Chitta Baral, Yezhou Yang% <-this % stops a space
    \IEEEcompsocitemizethanks{
        \IEEEcompsocthanksitem Corresponding authors: Tejas Gokhale (gokhale@umbc.edu) and Hamid Palangi (hpalangi@microsoft.com)
        \IEEEcompsocthanksitem T.G. is with the University of Maryland, Baltimore County.
        \IEEEcompsocthanksitem H.P., B.N., V.V., E.H., and E.C. are with Microsoft Research.
        \IEEEcompsocthanksitem C.B. and Y.Z. are with Arizona State University.
        \IEEEcompsocthanksitem Code: \href{https://github.com/microsoft/VISOR}{https://github.com/microsoft/VISOR}. 
        \IEEEcompsocthanksitem Data: \href{https://huggingface.co/datasets/tgokhale/sr2d_visor}{https://huggingface.co/datasets/tgokhale/sr2d\_visor}
    }
}

% The paper headers
\markboth{}{Gokhale \MakeLowercase{\textit{et al.}}: Benchmarking Spatial Reasoning Abilities of Text-to-Image Generative Models}
\IEEEpubid{}%0000--0000/00\$00.00~\copyright~2021 IEEE}
% \IEEEpubid{submission}
% Remember, if you use this you must call \IEEEpubidadjcol in the second
% column for its text to clear the IEEEpubid mark.

% % \maketitle
% \twocolumn[{%
% \renewcommand\twocolumn[1][]{#1}%
%   \maketitle
%   \centering
%   \vspace{-.3in}
%   \includegraphics[width=\linewidth]{figures/motivating_example_wide.pdf}
%   \vspace{-.1in}
%   \captionof{figure}{
%     We benchmark T2I models on their competency with generating appropriate spatial relationships in their visual renderings.
%     Although text inputs may explicitly mention these spatial relationships, we find that
%     T2I models lack such spatial understanding. 
%     Code: \href{https://github.com/microsoft/VISOR}{https://github.com/microsoft/VISOR}. Downloadable text and image data: \href{https://huggingface.co/datasets/tgokhale/sr2d_visor}{https://huggingface.co/datasets/tgokhale/sr2d\_visor}
%   }
%   \vspace{0.2in}
%   \label{fig:motivating_example}
% }]

\maketitle

\begin{abstract}
Spatial understanding is a fundamental aspect of computer vision and integral for human-level reasoning about images, making it an important component for grounded language understanding. 
While recent text-to-image synthesis (T2I) models have shown unprecedented improvements in photorealism, it is unclear whether they have reliable spatial understanding capabilities. 
We investigate the ability of T2I models to generate correct spatial relationships among objects and present VISOR, an evaluation metric that captures how accurately the spatial relationship described in text is generated in the image. 
To benchmark existing models, we introduce a dataset, \textit{SR\textsubscript{2D}}, that contains sentences describing two or more objects and the spatial relationships between them. 
We construct an automated evaluation pipeline to recognize objects and their spatial relationships, and employ it in a large-scale evaluation of T2I models.
Our experiments reveal a surprising finding that, although state-of-the-art T2I models exhibit high image quality, they are severely limited in their ability to generate multiple objects or the specified spatial relations between them.
Our analyses demonstrate several biases and artifacts of T2I models such as the difficulty with generating multiple objects, a bias towards generating the first object mentioned, spatially inconsistent outputs for equivalent relationships, and a correlation between object co-occurrence and spatial understanding capabilities. 
We conduct a human study that shows the alignment between VISOR and human judgement about spatial understanding.
We offer the SR\textsubscript{2D} dataset and the VISOR metric to the community in support of T2I reasoning research.

\end{abstract}


\section{Introduction}
Text to image synthesis (T2I), 
has advanced rapidly with capabilities for generating high-definition images in response to text prompts. 
Models are being used as tools for art, graphic design, and image editing. 
The power of T2I models for generating photorealistic objects and scenes is well-known. 
We less understand the ability of the models to faithfully render spatial relationships in its compositions.

\begin{figure*}
    \centering
    \includegraphics[width=\linewidth]{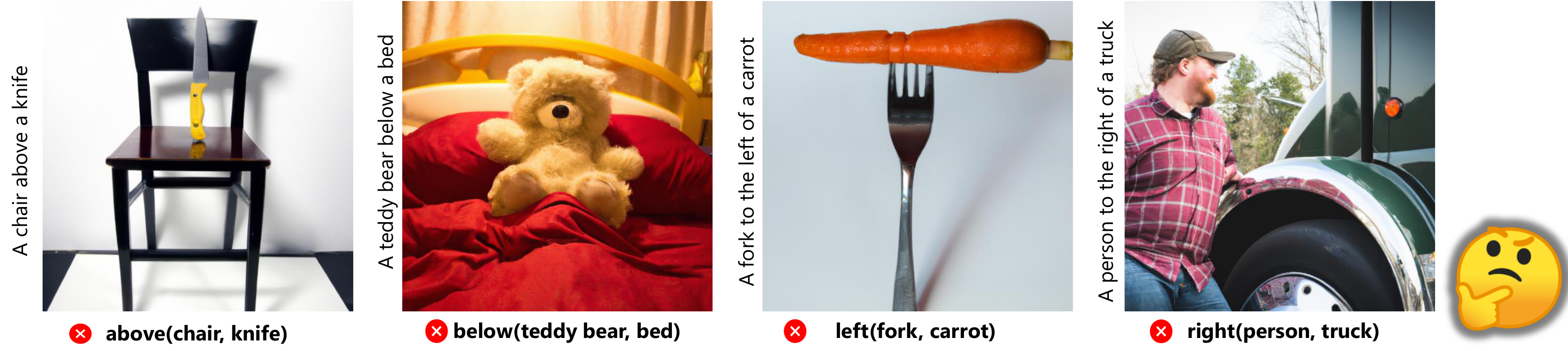}
    \caption{
    We benchmark T2I models on their competency with generating appropriate spatial relationships in their visual renderings.
    Although text inputs may explicitly mention these spatial relationships, we find that
    T2I models lack such spatial understanding. 
    % Code: \href{https://github.com/microsoft/VISOR}{https://github.com/microsoft/VISOR}. Downloadable text and image data: \href{https://huggingface.co/datasets/tgokhale/sr2d_visor}{https://huggingface.co/datasets/tgokhale/sr2d\_visor}
    }
    \label{fig:motivating_example}
\end{figure*}

We pursue the question: Do T2I models have the ability to render the spatial relationships among objects that are specified in text prompts?
\cref{fig:motivating_example} illustrates images generated by a state-of-the-art model (DALLE-v2~\cite{ramesh2022hierarchical}) for sentences that contain a spatial relationship between two objects.
In these examples, although both objects mentioned in the text are generated, the specified spatial relationship is not rendered. 

Spatial relations and larger scene geometries are integral aspects of computer vision. Rendering and reasoning about these relationships is crucial for many applications such as language-guided navigation and object manipulation~\cite{anderson2018vision,mees2020learning,nair2022learning}. A lack of spatial understanding by T2I models can be frustrating to creators seeking to render specific configurations of objects. The assertion of spatial relationship is common in natural communication among humans and poor capabilities in this realm will rapidly come to the fore in navigational and instructional applications.

Prior work on evaluation metrics for T2I models have focused on photorealism~\cite{salimans2016improved,heusel2017gans}, object accuracy~\cite{hinz2020semantic}, and image-text vector similarity (via CLIP~\cite{hessel2021clipscore}, retrieval~\cite{xu2018attngan}, and captioning~\cite{hong2018learning}).
We find that these metrics are insensitive to errors with generating spatial relationships (\cref{sec:existing}). This finding highlights the need for a metric to quantify competencies and progress in spatial understanding.
We develop an automated evaluation pipeline that employs computer vision to recognize objects and their spatial relationships, and harness this pipeline to conduct a large-scale evaluation of the spatial understanding capabilities of T2I models. We create the {``SR\textsubscript{2D}''} dataset, containing 25,280 sentences describing two-dimensional spatial relationships (\textit{left/right/above/below}) between pairs of commonly occurring objects from MS-COCO~\cite{lin2014microsoft}, as shown in \Cref{tab:examples_dataset}.
We study several state-of-the-art models: GLIDE~\cite{nichol2021glide}, DALLE-mini~\cite{dayma2021dallemini}, CogView2~\cite{ding2022cogview2}, DALLE-v2~\cite{ramesh2022hierarchical}, Stable Diffusion~\cite{rombach2022high}, and Composable Diffusion~\cite{liu2022compositional}.
For each model we generate and evaluate four images per SR\textsubscript{2D} example, i.e., a large-scale study of 101,120 images per model.
Our study makes significant advances to evaluation of T2I reasoning capabilities since we evaluate photorealistic images rather than synthetic objects on solid background.

We introduce a new evaluation metric we refer to as \textsc{VISOR} (for \textbf{v}erify\textbf{i}ng \textbf{s}patial \textbf{o}bject \textbf{r}elationships), 
to compare the spatial understanding abilities of T2I models. We define three variants of the metric:
(1) VISOR: verifies spatial correctness for each image w.r.t.\ its text input, 
(2) VISOR\textsubscript{n}: consider whether at least $n$ of the multiple generated images for each text input are spatially correct, 
(3) VISOR\textsubscript{cond}: verifies the spatial correctness in images, conditioned on both objects being generated by the model.
While VISOR provides a macro-perspective on the performance gap in the spatial capabilities of T2I models, VISOR\textsubscript{n} reflects the practical value of the model to users who can select one of many images generated by the model.
The conditional formulation VISOR\textsubscript{cond} disentangles two capabilities: (i) the generation of multiple objects and (ii) generation of correct spatial relationships between the rendered objects.
We conduct a human study on Amazon Mechanical Turk and find that the VISOR metric is correlated with human judgment.

Our experiments reveal several interesting findings.
First, we find that all existing models are significantly worse at generating two objects as compared to their capability to render single objects.
While previous work shows exceptional zero-shot compositionality of colors, styles, and attributes \cite{ramesh2022hierarchical,saharia2022photorealistic,yu2022scaling},
we found challenges with compositionality for multiple objects.
Second, we find poor spatial understanding: even in cases where both objects are generated, models tend to ignore spatial relationships specified in language.
VISOR scores for all models show that even the best model in our benchmark generates correct spatial relationships on less than $40\%$ of test cases.
When we consider a strict metric (VISOR\textsubscript{4}) that requires that all generated images for text prompts to have correct spatial relationships, the best model (DALLE-v2) achieves the goal in $7.49\%$ cases.
Third, we discover several biases in T2I models: 
to generate only the first object mentioned in the text and ignoring the second, to show better performance on commonly occurring object pairs, to have a tendency to merge two objects into one, and to have inconsistent outputs for equivalent text inputs.

\begin{table}[t]
    \centering
    \resizebox{\linewidth}{!}{
    \begin{tabular}{@{}llll@{}}
        \toprule
        \textbf{A} & \textbf{B} & \textbf{R} & \textbf{Text}\\
        \midrule
        microwave   & sink      & left  & A microwave to the left of a sink \\
        elephant    & cat       & right & An elephant to the right of a cat \\ 
        donut       & airplane  & above & A donut above an airplane \\
        suitcase    & chair     & below & A suitcase below a chair \\
        keyboard    & bench     & left  & A keyboard to the left of a bench \\
        bed         & bear      & right & A bed to the right of a bear \\
        potted plant & fire hydrant & above & A potted plant above a fire hydrant \\
        person      & umbrella  & below & A person below an umbrella \\
        \bottomrule
    \end{tabular}
    }
    \caption{Examples of text inputs from the SR\textsubscript{2D} dataset for a pair of objects (A, B) and relationship R between them.}
    \label{tab:examples_dataset}
\end{table}

\noindent To summarize, our contributions are as follows:
\begin{itemize}[noitemsep]
    \item We introduce a metric called VISOR to quantify spatial reasoning performance.  VISOR can be used off-the-shelf with any text-to-image model, disentangles correctness of object generation with the ability of spatial understanding.
    \item We construct and make available a large-scale dataset: SR\textsubscript{2D}, which contains sentences that describe spatial relationships between a pair of 80 commonly occurring objects along with linguistic variations.
    \item With SR\textsubscript{2D}, we conduct a large-scale benchmarking of state-of-the-art T2I models with automated and human evaluation of spatial reasoning abilities of state-of-the-art T2I models using the VISOR metric.  We find that although existing T2I models have improved photorealism, they lack spatial and relational understanding with multiple objects, and indicate several biases.
\end{itemize}

%%%%%%%%%%%%%%%%%%%%%%%%%%%%%%%%%%%%%%%%%%%%%%%%
\section{Related Work}
\subsection{Text-to-Image Synthesis.} 
Earlier work~\cite{reed2016generative,zhang2017stackgan} trained and evaluated models on human-labeled datasets~\cite{welinder2010caltech,nilsback2008automated,lin2014microsoft}.
Recent work on T2I has focused on zero-shot capabilities by taking advantage of implicit knowledge from pretrained language models and V+L models like CLIP, and the diffusion technique to train on large-scale web data.

\subsection{Biases in Vision+Language models} have been studied from a linguistic perspective, such as question-answer priors in VQA~\cite{agrawal2018don,kervadec2020roses}, gender bias in captioning~\cite{hendricks2018women,zhao2017men}, shortcut effects in commonsense reasoning~\cite{ye2021case}, and failure modes in logic-based VQA~\cite{ray2019sunny,gokhale2020vqa,goel2020iq}.
The difficulty of spatial understanding has been studied for visual grounding~\cite{liu2019clevr}, image-text matching~\cite{liu2022visual}, VQA~\cite{johnson2017clevr,hudson2018compositional}, and navigation~\cite{chen2019touchdown}.

\subsection{Human Study about Relational Understanding.}
Conwell \etal~\cite{conwell2022testing} conducted a human study (1350 images) of DALLE-v2 on a set of eight physical relations and seven action-based relations between 12 object categories.
Our human study is significantly larger in scale, considers diverse text inputs, several state of the art models, and establishes an alignment with the automated VISOR metric.

\subsection{Empirical Evaluation of Visual Reasoning Skills.}
DALL-Eval~\cite{cho2022dall} evaluates reasoning skills of T2I models trained and tested on a synthetically generated dataset \textsc{PaintSkills} with black backgrounds and 21 rendered object categories. In our work, we instead focus on the evaluation of photorealistic and open-domain images with commonly occurring real-world objects and backgrounds on a large scale. Most importantly, we devise a new human-aligned metric (VISOR) that disentangles object accuracy from spatial understanding to accordingly measure progress in spatial reasoning despite the model's capabilities in object generation.

\subsection{Other Failure Modes of T2I Models.}
Preliminary stress-testing of DALLE-v2~\cite{marcus2022very} (14 prompts), \cite{leivada2022dall} (40 prompts) and \cite{saharia2022photorealistic} (200 prompts) illustrated anecdotal failures of the model in terms of compositionality, grammar, binding, and negation.
However, since these studies rely on human judgment, there is a need for automated evaluation techniques for comparing the reasoning abilities of T2I models.
Our paper fills this gap with the automated VISOR metric for spatial relationships and the large-scale SR\textsubscript{2D} dataset.

%%%%%%%%%%%%%%%%%%%%%%%%%%%%
\section{Spatial Relationships Challenge Dataset} \label{sec:sr2d}

\subsection{Predicate Generation.}
Our goal is to collect a set of sentences that describe spatial relationships between two objects.
Let $\mathcal{C}$ be the set of object categories.
Let $\mathcal{R}$ be the set of spatial relationships between objects.
In this paper, we focus on two-dimensional relationships, i.e. 
$\mathcal{R} = \{ left, right, above, below\}$,
and $80$ object categories derived from the MS-COCO dataset~\cite{lin2014microsoft}.
Then, for every $A\in\mathcal{C}$, $B\in\mathcal{C}$, and $R\in\mathcal{R}$, let the predicate $R(A, B)$ indicate that the spatial relationship $R$ exists between object $A$ and object $B$. 
For example $\relleft(\texttt{cat}, \texttt{dog})$ describes a scene where a cat is to the left of a dog.
For each pair, we construct 8 types of spatial relationships as shown below:\\

\setlength{\fboxsep}{0pt}
\noindent\fbox{%
    \parbox{\linewidth}{
        \centering
        \smallskip
        % \small
        \it
        left(A, B), right(A, B), above(A, B), below(A, B)\\
        left(B, A), right(B, A), above(B, A), below(B, A)
        \smallskip
    }
}

\subsection{Sentence Generation.}
For each predicate $R(A, B)$, we convert it into a template \fbox{\strut\texttt{ <A> <R> <B> }} and paraphrase it into natural language.
Appropriate articles \textit{``a''/``an''} are prepended to object names \texttt{A} and \texttt{B}, to obtain four templates:

\smallskip
\setlength{\fboxsep}{0pt}
\noindent\fbox{%
    \parbox{\linewidth}{
    % \small
    \smallskip
    \centering 
    A/an \texttt{<A>} \textit{to the left of} a/an \texttt{<B>}\\
    A/an \texttt{<A>} \textit{to the right of} a/an \texttt{<B>}\\
    A/an \texttt{<A>} \textit{above} a/an \texttt{<B>}\\
    A/an \texttt{<A>} \textit{below} a/an \texttt{<B>}
    \smallskip
}
}

\smallskip
The template-based procedure has several advantages.
First, it avoids linguistic ambiguity, subjectivity, and grammatical errors.
Second, it is extensible to new object categories and additional spatial relationships. 
While we focus on two-dimensional relationships in this paper, our templates can be extended for generating test inputs for studying more complex spatial relationships and geometric features of objects, as we discuss in \cref{sec:discussion}.

\subsection{Dataset Statistics.}
We use $|\mathcal{C}|=80$ object categories from MS-COCO and  therefore obtain ${80 \choose 2} = 3160$ unique combinations of object pairs (A, B).
For each pair, we construct 8 types of spatial relationships listed above, which leads to a total of 3,160$\times$ 8 $=$ 25,280 predicates.
The SR\textsubscript{2D} dataset contains 25,280 text prompts, uniformly distributed across 80 COCO object categories, with each object being found in 632 prompts.
\cref{tab:examples_dataset} lists a few illustrative examples.

%%%%%%%%%%%%%%%%%%%%%%%%%%%\SEC 3 %%%%%%%%%%%%%%%%%%%%%%%%%%%
\begin{figure}[t]
    \centering
    \includegraphics[width=\linewidth]{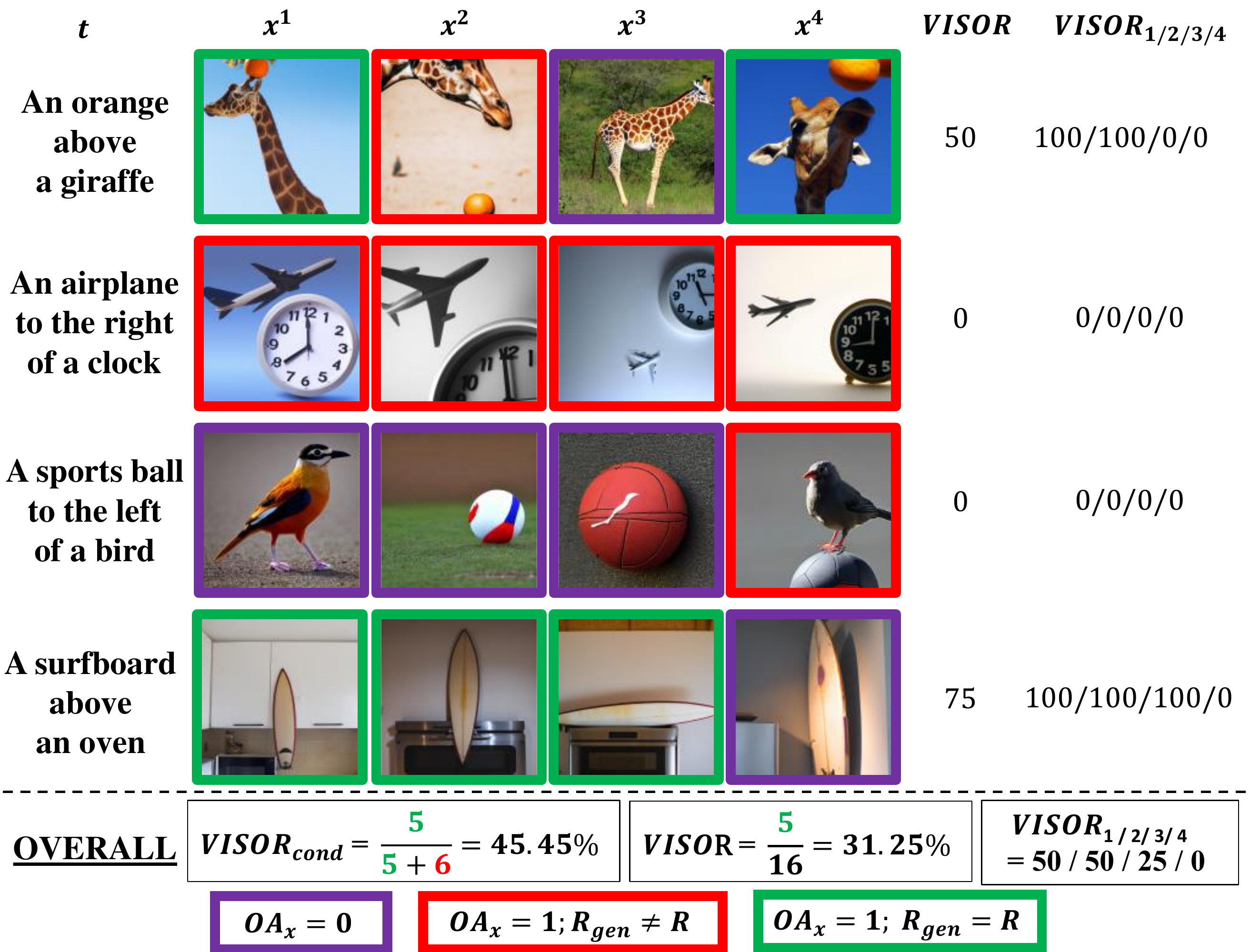}
    \caption{
    Examples illustrating the intuition behind OA, VISOR, VISOR\textsubscript{cond}, and VISOR\textsubscript{1/2/3/4}. 
    {\color{darkpurple}\bf Purple box}: cases where one or both objects are not generated; {\color{red}\bf Red box}: both objects are generated but with a wrong spatial relationship; {\color{darkgreen}\bf Green box}: successful cases. 
    }
    \label{fig:VISOR_example}
\end{figure}

\begin{figure*}[t]
    \centering
    \includegraphics[width=\linewidth]{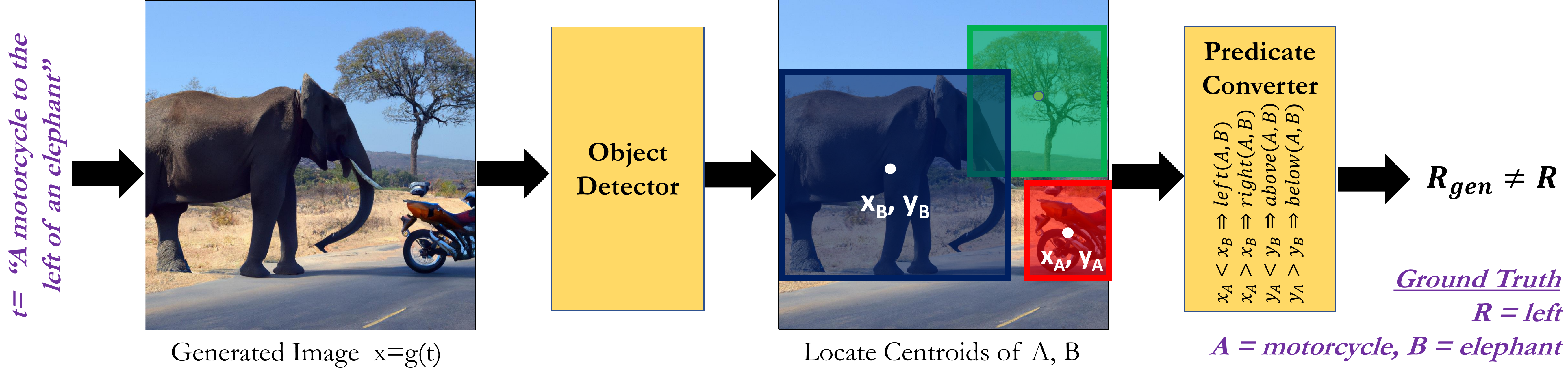}
    \caption{
    For text $t$ and corresponding generated image $x=g(t)$, object centroids are located and converted into predicates indicating the spatial relationship between them. These predicates are compared with the ground truth relationship $R$ to obtain the VISOR score.}
    \label{fig:VISOR_pipeline}
\end{figure*}

 %%%%%%%%%%%%%%%%%%%%%%%%%
\section{VISOR Evaluation Metric}\label{sec:VISOR}
We propose VISOR as an automated metric for quantifying spatial understanding abilities of text-to-image models.
VISOR is short for \textbf{V}er\textbf{i}fying \textbf{S}patial \textbf{O}bject \textbf{R}elationships.

\subsection{Definitions}
\begin{definition}[Object Accuracy]\label{def:oa_x}
    Let $h$ be an oracle function that returns a set of detected objects in image $x$ from set $\mathcal{C}$. Then, object accuracy for an image $x$, generated by a sentence containing objects $A$ and $B$ is:
    \begin{equation}
        % \small
        \textsc{OA}(x, A, B) = \mathbbm{1}_{h(x)}(\exists A \cap \exists B).
        \label{eq:oa_x}
    \end{equation}
\end{definition}
Note that, the oracle function $h$ here could be either a pluggable learned model or a human detecting the presence of objects mentioned in the sentence. In our experiments, we show results for both cases and a correlational analyses between the two. 
Object accuracy is agnostic to the relationship $R$, whose presence is instead captured in the VISOR metric. 

\begin{definition}[VISOR]\label{def:VISOR_uncond}
Let $R_{gen}$ be the generated spatial relationship, while $R$ is the ground-truth relationship mentioned in text.
Then, for each image $x$, 

\begin{equation}
    % \small
    \textsc{VISOR} (x, A, B, R) = 
    \begin{cases}
    1, ~ \text{if } (R_{gen}=R) \cap \exists A \cap \exists B\\
    0, ~ \text{otherwise}.
    \end{cases}
    \label{eq:VISOR_uncond}
\end{equation}
\end{definition}

A useful feature of T2I models for artists and designers is the ability to generate multiple images for each input text prompt. 
This allows the creators to pick an appropriate image from $N$ generated images.
We define VISOR\textsubscript{n} to reflect how good T2I models are at generating at least $n$ spatially correct images given a text input that mentions a spatial relationship. From a usability perspective (where creators have the option to pick from the output image set), VISOR\textsubscript{n}
is useful for measuring if it is possible to find at least $n$ images that  satisfy the prompt.
\begin{definition}[VISOR\textsubscript{n}]
    \textsc{VISOR}\textsubscript{n} is the probability of generating images such that for every text prompt $t$, at least $n$ out of $N$ images have \textsc{VISOR}$=$1:
\begin{equation}
    \textsc{VISOR}_n (x,A,B,R) =
    \begin{cases}
	1,~ \textstyle\sum\limits_{i=1}^{N}\textsc{VISOR}(x_i, A, B,R) \geq n\\
        0,~ \text{otherwise}.
    \end{cases}
    \label{eq:VISOR_n}
\end{equation}
\end{definition}
        % \begin{numcases}{\textsc{VISOR}_n (x,A,B,R) =}
        %     
        % \end{numcases}

\noindent The relationship between VISOR and VISOR\textsubscript{n} is given below. The proof is presented in the supplementary materials.
\begin{equation}\label{eq:visor_proof}
    % \small
    \textsc{VISOR} = \frac{1}{N} \sum_{n=0}^{N} n (\textsc{VISOR}_n - \textsc{VISOR}_{n+1}).
\end{equation}
In our study we use $N=4$ images per text prompt and, therefore, report VISOR\textsubscript{1}, VISOR\textsubscript{2}, VISOR\textsubscript{3}, and VISOR\textsubscript{4}.
\cref{fig:VISOR_example} shows an example computation of all VISOR metrics.

Note that $\textsc{VISOR}=1$ only if both objects are generated in the image, i.e. $\textsc{OA}=1$.
However, as we will see in \cref{sec:experiments}, T2I models fail to generate multiple objects in a large subset of images.
As such, it is important to disentangle the two abilities of the models to (1) generate multiple objects and (2) to generate them according to the spatial relationships described in the text of the prompt.
For this purpose, we define conditional VISOR:
\begin{definition}[Conditional VISOR]\label{def:VISOR_av}
    is defined as the conditional probability of correct spatial relationships being generated, given that both objects were generated correctly.
    \begin{equation}
        \label{eq:VISOR_cond}
        % \small
        \textsc{VISOR}_{cond}   = P(R_{gen}{=}R | \exists A \cap \exists B)
    \end{equation}
\end{definition}

\subsection{Implementation}
The VISOR computation process is summarized in \cref{fig:VISOR_pipeline}.
Given any text prompt $t$ and a T2I model $g$, we first generate images $x=g(t)$, and use an object detector to localize objects in $x$.
Object accuracy OA is computed using \cref{eq:oa_x}.
We obtain centroid coordinates of objects A and B from the the bounding boxes of the detected objects.
Based on the centroids, we deduce the spatial relationship $R_{gen}$ between them using the rules shown in the ``Predicate Converter'' box in \cref{fig:VISOR_pipeline}.
Finally, the generated relationship is compared with the ground-truth relationship $R$, and VISOR scores are computed using \cref{eq:VISOR_uncond,eq:VISOR_n,eq:VISOR_cond}.

We use OWL-ViT~\cite{minderer2022simple}, a state of the art open-vocabulary object detector, with a CLIP backbone and ViT-B/32 transformer architecture and confidence threshold $0.1$.
The supplementary material also contains results using DETR-ResNet-50~\cite{carion2020end} trained on MS-COCO.
The results using both object detectors are similar and lead to an identical ranking of models in our benchmark.
However, the open-vocabulary functionality of OWL-ViT ensures that VISOR is widely applicable to other datasets, categories, and vocabularies.
This removes dependence on specific datasets, making VISOR widely applicable for any freeform text input.

%%%%%%%%%%%%%%%%%%%%%%%%%%%%%%%%%%
\begin{table*}[t]
    \centering
    \large
    \resizebox{\linewidth}{!}{
    \begin{tabular}{@{}l rrrr rrrr r r @{}}
        \toprule
        {\textbf{Model}} & \textbf{BLEU-1} & \textbf{BLEU-2} & \textbf{BLEU-3} & \textbf{BLEU-4} & \textbf{METEOR} & \textbf{ROUGE} & \textbf{CIDER} &  \textbf{SPICE} & \textbf{CLIPScore} & \textbf{VISOR}\\
        \midrule
        GLIDE       & 0.29 / {\color{magenta}0} & 0.14 / {\color{magenta}1.9e-6} & 0.05 / {\color{magenta}2.9e-6} & 0.02 / {\color{magenta}7.4e-7} & 0.13 / {\color{magenta}-9.9e-6} & 0.35 / {\color{magenta}-2.6e-6} & 0.18 / {\color{magenta}1.1e-6} & 0.11 / {\color{magenta}-8.1e-6} & 0.70 / {\color{magenta}1.3e-3} & 0.02 / {\color{darkgreen}0.03} \\
        GLIDE + CDM & 0.31 / {\color{magenta}0} & 0.15 / {\color{magenta}2.9e-5} & 0.06 / {\color{magenta}7.6e-6} & 0.02 / {\color{magenta}1.7e-6} & 0.15 / {\color{magenta} 6.4e-5} & 0.36 / {\color{magenta}1.7e-6} & 0.22 / {\color{magenta}3.1e-5} & 0.14 / {\color{magenta}-7.7e-5} & 0.75 / {\color{magenta}-8.8e-6} & 0.06 / {\color{darkgreen}0.07} \\
        DALLE-mini  & 0.34 / {\color{magenta}0} & 0.19 / {\color{magenta}-4.4e-5} & 0.09 / {\color{magenta}-1.6e-5} & 0.04 / {\color{magenta}-4.5e-6} & 0.19 / {\color{magenta}2.1e-6} & 0.41 / {\color{magenta}-7.3e-6} & 0.34 / {\color{magenta}-6.0e-5} & 0.20 / {\color{magenta}3.3e-5} & 0.80 / {\color{magenta}1.5e-3} & 0.16 / {\color{darkgreen}0.22} \\
        CogView-2   & 0.30 / {\color{magenta}0} & 0.16 / {\color{magenta} 4.4e-6} & 0.07 / {\color{magenta}1.3e-6} & 0.03 / {\color{magenta}5.6e-6} & 0.16 / {\color{magenta}-6.6e-6} & 0.36 / {\color{magenta}-2.9e-6} & 0.25 / {\color{magenta}8.7e-6} & 0.15 / {\color{magenta}7.0e-5} & 0.72 / {\color{magenta}1.5e-5} & 0.12 / {\color{darkgreen}0.13}\\
        DALLE-v2    & 0.36 / {\color{magenta}0} & 0.21 / {\color{magenta}-1.9e-5} & 0.11 / {\color{magenta}-4.8e-6} & 0.04 / {\color{magenta}-1.5e-6} & 0.21 / {\color{magenta}1.7e-4} & 0.44 / {\color{magenta}8.8e-6} & 0.40 / {\color{magenta}-2.8e-5} & 0.22 / {\color{magenta}-4.1e-5} & 0.84 / {\color{magenta}1.8e-3} & 0.38 / {\color{darkgreen}0.55} \\
        SD          & 0.33 / {\color{magenta}0} & 0.18 / {\color{magenta}3.3e-6} & 0.08 / {\color{magenta}9.7e-7} & 0.03 / {\color{magenta}2.8e-7} & 0.19 / {\color{magenta}1.0e-5} & 0.40 / {\color{magenta}-2.6e-6} & 0.31 / {\color{magenta}4.3e-6} & 0.19 / {\color{magenta}7.4e-5} & 0.79 / {\color{magenta}1.5e-3} & 0.19 / {\color{darkgreen}0.23} \\
        SD + CDM    & 0.32 / {\color{magenta}0} & 0.17 / {\color{magenta}1.1e-5} & 0.07 / {\color{magenta}5.1e-6} & 0.03 / {\color{magenta}1.3e-6} & 0.17 / {\color{magenta}1.6e-4} & 0.38 / {\color{magenta}4.4e-6} & 0.28 / {\color{magenta}1.2e-5} & 0.18 / {\color{magenta}-4.5e-5} & 0.77 / {\color{magenta}3.6e-4} & 0.15 / {\color{darkgreen}0.17}\\
        SD 2.1  & 0.35 / {\color{magenta}0} & 0.20 / {\color{magenta}-1.3e-5} & 0.09 / {\color{magenta}4.2e-6} & 0.038 / {\color{magenta}-1.3e-6} & 0.20 / {\color{magenta}7.1e-5} & 0.42 / {\color{magenta}5.4e-6} & 0.35 / {\color{magenta}-1.8e-5} & 0.20 / {\color{magenta}3.5e-5} & 0.82 / {\color{magenta}1.0e-3} & 0.30 / {\color{darkgreen}0.37} \\
        \bottomrule
    \end{tabular}
    }
    % \vspace{0.5pt}
    \caption{
        $s/\Delta_s$ scores for T2I metrics shown in the 0 to 1 range.
        All previous metrics have low $\Delta_s$ ({\color{magenta}magenta}) whereas VISOR has high $\Delta_s$ ({\color{darkgreen}green}), showing they are ineffective in quantifying and benchmarking spatial understanding.
    }
    \label{tab:delta_caption}
\end{table*}

\begin{table*}[t]
    \centering
    % \LARGE
    % \resizebox{\linewidth}{!}{
    \begin{tabular}{@{}l rr r cccc@{}}
        \toprule
         & \multirow{2}{*}{\textbf{OA (\%)}} & \multicolumn{6}{c}{\textbf{VISOR (\%)}} \\
         \cmidrule{3-8}
        \textbf{Model} & & \textbf{uncond} & \textbf{{cond}} & \textbf{1} & \textbf{2} & \textbf{3} & \textbf{4}\\
        \midrule
        GLIDE \cite{nichol2021glide}           &  3.36 &  1.98 & 59.06 &  6.72 & 1.02 & 0.17 & 0.03\\ 
        GLIDE + CDM \cite{liu2022compositional}    & 10.17 &  6.43 & 63.21 & 20.07 &  4.69 &  0.83 & 0.11 \\
        DALLE-mini \cite{dayma2021dallemini}    & 27.10 & 16.17 & 59.67 & 38.31 & 17.50 &  6.89 & 1.96 \\
        CogView2 \cite{ding2022cogview2}       & 18.47 & 12.17 & \textbf{65.89} & 33.47 & 11.43 &  3.22 & 0.57 \\
        DALLE-v2 \cite{ramesh2022hierarchical}       & \textbf{63.93} & \textbf{37.89} & 59.27 & \textbf{73.59} & \textbf{47.23} & \textbf{23.26} & \textbf{7.49} \\
        SD \cite{rombach2022high}             & 29.86 & 18.81 & 62.98 & 46.60 & 20.11 &  6.89 & 1.63 \\
        SD + CDM \cite{liu2022compositional}      & 23.27 & 14.99 & 64.41 & 39.44 & 14.56 &  4.84 & 1.12\\
        SD 2.1      & 47.83 & 30.25 & 63.24 & 64.42 & 35.74 & 16.13 &  4.70 \\  
        Structured Diffusion \cite{feng2022training}    & 28.65 & 17.87 & 62.36 & 44.70 & 18.73 & 6.57 & 1.46 \\
        Attend-and-Excite \cite{chefer2023attendandexcite}       & 42.07 & 25.75 & 61.21 & 49.29 & 19.33 & 4.56 & 0.08\\
        \bottomrule
    \end{tabular}
    % }
    \caption{Comparison of the performance of all models in terms of object accuracy (OA) and each version of VISOR.}
    \label{tab:visor_main}
\end{table*}

%%%%%%%%%%%%%%%%%%%%%%%%%%%%%%%%%%%%%%%%%%%
\section{Experiments}\label{sec:experiments}
% \noindent\textbf{Baselines.}
In the following experiments, we study state-of-the-art T2I models as baselines, including
GLIDE~\cite{nichol2021glide}, DALLE-mini~\cite{dayma2021dallemini}, CogView2~\cite{ding2022cogview2}, DALLE-v2~\cite{ramesh2022hierarchical}, and Stable-Diffusion (SD and SD 2.1.)~\cite{rombach2022high}, and two versions of Composable Diffusion Models~\cite{liu2022compositional} (GLIDE + CDM and SD + CDM).
We generate $N{=}4$ images for each text prompt from our SR\textsubscript{2D} dataset, to obtain 126,720 images per model and compare performance in terms of OA, VISOR, VISOR\textsubscript{cond}, and VISOR\textsubscript{1/2/3/4}.
We also included two models that use additional knowledge for text-to-image generation. 
These are: Structured Diffusion \cite{feng2022training} which uses constituency parsing and tree structures from linguistics, and Attend-Excite \cite{chefer2023attendandexcite} which requires human  users to input the indices of the object tokens.

\subsection{Ineffectiveness of Existing Metrics}
\label{sec:existing}
T2I models have been primarily compared in terms of photorealism (purely visual) and human judgment about image quality (subjective).
We quantify whether existing automated multimodal metrics are useful for evaluating spatial relationships generated by T2I models.
We consider CLIPScore~\cite{hessel2021clipscore} (cosine similarity between image and text embeddings) and image captioning-based evaluation (BLEU~\cite{papineni2002bleu}, METEOR~\cite{banerjee2005meteor}, ROUGE~\cite{lin2004rouge}, CIDER~\cite{vedantam2015cider}, SPICE~\cite{anderson2016spice}) which are used by generating a caption $c$ for the synthesized image $x=g(t)$ and computing the captioning score with respect to the reference input text $t$.
Note that purely visual metrics (FID and Inception Score~\cite{heusel2017gans,salimans2016improved}) ignore the text, while semantic object accuracy~\cite{hinz2020semantic} ignores all words except nouns, making them incapable of scoring spatial relationships.

Let $s^t$ be the score for $(x, t)$ where $x$ is the generated image and $t$ is the input text.
Let $t_{flip}$ be the transformed version of $t$ obtained by inverting/flipping the spatial relationship in $t$ (for example, left$\rightarrow$right).
Let $s_{flip}^t$ be the score for $(x, t_{flip})$.
For each metric, we define $\Delta_s$ as the average difference between $s^t$ and $s^t_{flip}$ over the entire SR\textsubscript{2D} dataset:
\begin{equation}
    \Delta_s = \mathbb{E}_t [s^t - s_{flip}^t], 
\end{equation}

Thus, $\Delta_s$ captures the ability of metric $s$ to understand spatial relationships.
\Cref{tab:delta_caption} shows $s$ and $\Delta_s$ values for each previous metric and VISOR for each model.
It can be seen that, for all previous metrics, $\Delta_s$ is negligible and close to zero, which implies that they return similar scores even if the text is flipped.
For some cases, the difference is negative, implying that the score for the image and the flipped caption is higher.
On the other hand, the $\Delta$ values for VISOR are high implying that VISOR assigns significantly lower scores for the flipped samples.
These results establish the need for a new evaluation metric since none of the existing metrics are able to quantify spatial relationships reliably, and show the efficacy of VISOR for this purpose.

\subsection{Benchmarking Results}
\Cref{tab:visor_main} shows the results of benchmarking on our SR\textsubscript{2D} dataset.
We first note that the object accuracy of all models except DALLE-v2 is lower than 30\%.
While DALLE-v2 (63.93\%) significantly outperforms other models, it still shows a large number of failures in generating both objects that are mentioned in the prompt.
For the unconditional metrics VISOR and VISOR\textsubscript{1/2/3/4}, DALLE-v2 is the best performing model.
However, in terms of VISOR\textsubscript{cond}, CogView2 has the highest performance. This implies that, although CogView2 is better than other models on those examples where both objects are generated, the large failures of CogView2 in OA result in a lower unconditional VISOR score.
VISOR\textsubscript{4} is extremely low for all models including DALLE-v2 ($8.54\%$), revealing a large gap in performance.

\begin{figure}
    \centering
    \includegraphics[width=\linewidth]{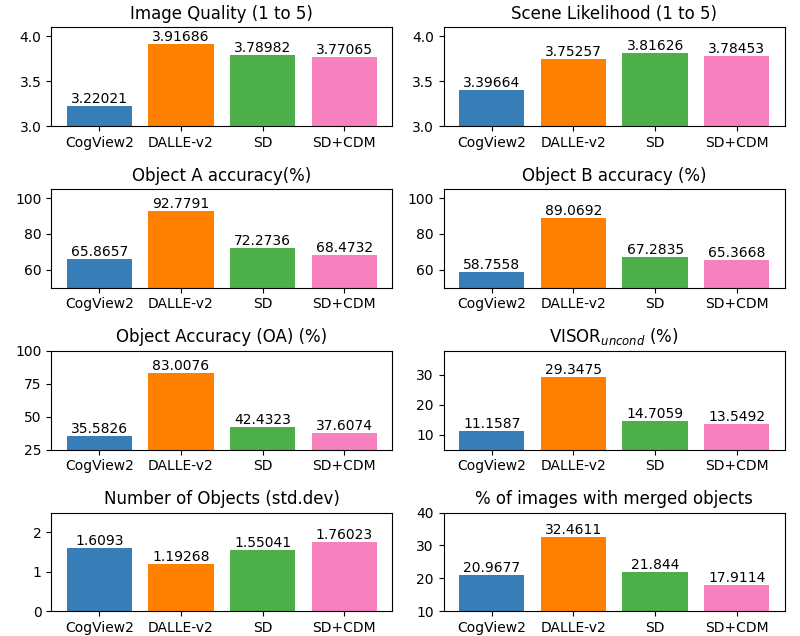}
    \caption{Summary of responses to each question in the human study, compared across all four models.}
    \label{fig:humanstudy_mean}
\end{figure}
\begin{figure*}
    \centering
    \fbox{
    \includegraphics[width=0.9\linewidth]{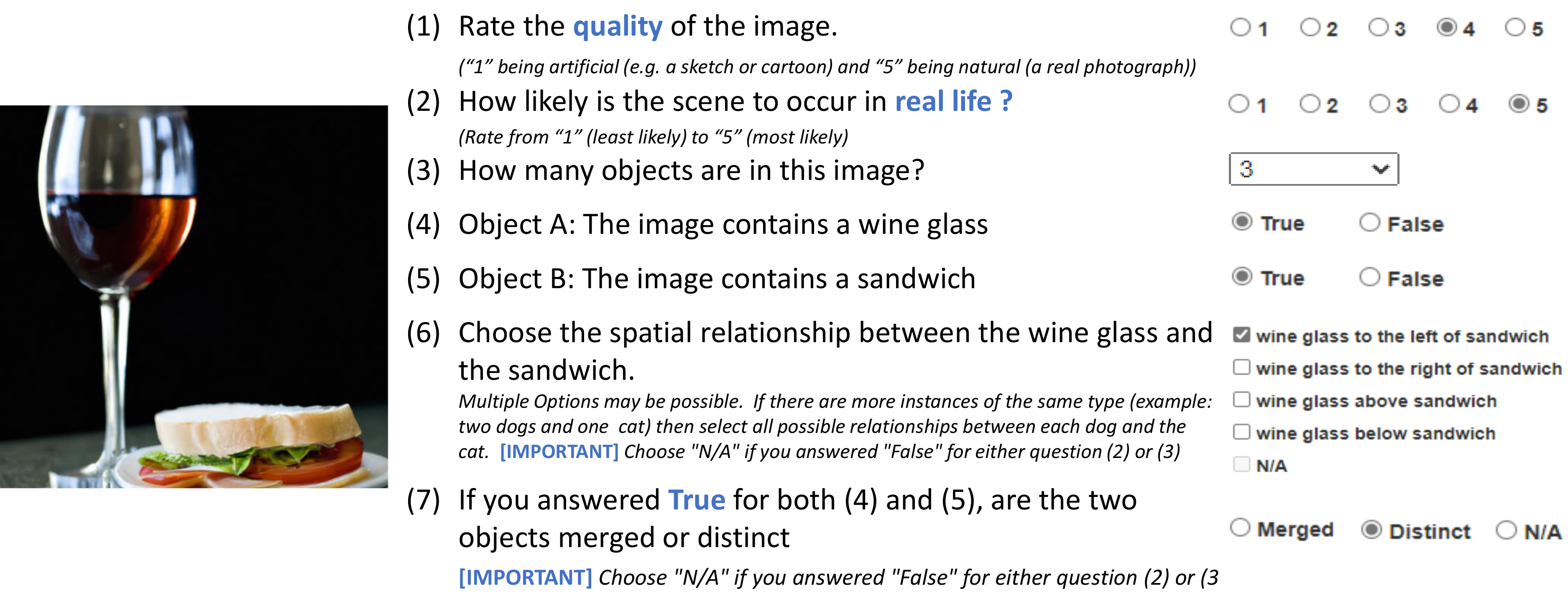}
    }
    \caption{The human study interface with an image on the left and seven multiple choice questions about it.}
    \label{fig:humanstudy_layout_image}
\end{figure*}

\subsection{Human Study}
\noindent\textbf{Methodology.}
We conducted a human evaluation study to understand the alignment of our metrics with human judgment, and to quantify the gap between object detector performance and human assessments of object presence.
For the human study, we used four models: CogView2, DALLE-v2, Stable Diffusion (SD), and SD + CDM.
Annotators were shown (via Amazon Mechanical Turk) an image generated by one of the four models, and were asked seven questions about it, as shown in \cref{fig:humanstudy_layout_image}. The questions assessed human evaluation of image quality and scene realism (\emph{scene likelihood}) on a Likert scale (1 through 5), the number of objects, answering True or False for presence of objects, selecting valid spatial relationships, and responding if two objects were merged in the image.
We used a sample size of 1000 images per model and 3 workers per sample.

\medskip\noindent\textbf{Results.}
\cref{fig:humanstudy_mean} shows a summary of responses for each question in the human study.
While DALLE-v2 received the highest image quality rating, SD and SD+CDM received higher scene likelihood rating.
Interestingly, DALLE-v2 also had the largest number of images with merged objects (32.46\%); several cases of this phenomenon are shown in \cref{fig:merged}.
Inter-annotator agreement was high for all questions in terms of majority (agreement between at least 2 out of 3 workers) and unanimous agreement (agreement between all 3 out of 3 workers) as reported in \Cref{tab:mturk_agreement}.

\begin{table*}
    \centering
    % \large
    % \resizebox{\linewidth}{!}{
    \begin{tabular}{l c c c c}
        \toprule
        \textbf{Response} & CogView2 & DALLE-v2 & SD & SD + CDM \\
        \midrule
        Image Quality       & 65.47 / 52.93 & 75.02 / 62.33 & 69.86 / 55.31 & 72.59 / 57.99 \\
        Scene Likelihood    & 64.40 / 50.78 & 72.13 / 59.62 & 69.47 / 52.35 & 67.19 / 53.99 \\
        Num. Objects        & 79.63 / 50.03 & 87.09 / 46.39 & 81.41 / 46.06 & 80.28 / 45.74 \\
        Object A            & 100.0 / 33.00 & 99.64 / 8.02  & 100.0 / 18.56 & 100.0 / 20.04 \\
        Object B            & 100.0 / 32.75 & 100.0 / 13.39 & 100.0 / 22.44 & 100.0 / 25.51 \\
        Spatial Relation    & 100.0 / 23.33 & 100.0 / 47.90 & 100.0 / 30.79 & 100.0 / 25.00 \\
        Merged/Distinct     & 100.0 / 43.02 & 99.64 / 58.85 & 100.0 / 39.95 & 100.0 / 38.60 \\
        \bottomrule
    \end{tabular}
    % }
    \caption{Majority / Unanimous inter-worker agreement (\%) for each question in our human study.}
    \label{tab:mturk_agreement}
\end{table*}
\begin{table}[t]
    \centering
    \begin{tabular}{lcccc}
        \toprule
        \textbf{Metric} & CogView2 & DALLE-v2 & SD & SD-CDM \\
        \midrule
        OA                          & 73.07 & 73.87 & 79.25 & 80.21 \\
        VISOR\textsubscript{uncond} & 88.48 & 77.41 & 88.43 & 88.80 \\
        VISOR\textsubscript{cond}   & 75.02 & 75.62 & 76.95 & 74.69 \\
        \bottomrule
    \end{tabular}
    \caption{Agreement(\%) of human responses with automated metrics}
    \label{tab:human_visor_agreement}
\end{table}

\medskip\noindent\textbf{Alignment of VISOR with Human Responses.}
We observe that the ranking of models in terms of both object accuracy (OA) and VISOR is identical for the human study and for the automated VISOR scores in \Cref{tab:visor_main}, i.e. \textit{DALLE-v2 $>$ SD $>$ SD-CDM $>$ CogView2}.
\Cref{tab:human_visor_agreement} shows the percentage of samples for which responses from humans matched our automated evaluation using object detectors.

%%%% ANALYSIS %%%%
\section{Analysis}\label{sec:analysis}
\begin{figure}[t]
    \centering
    \includegraphics[width=\linewidth]{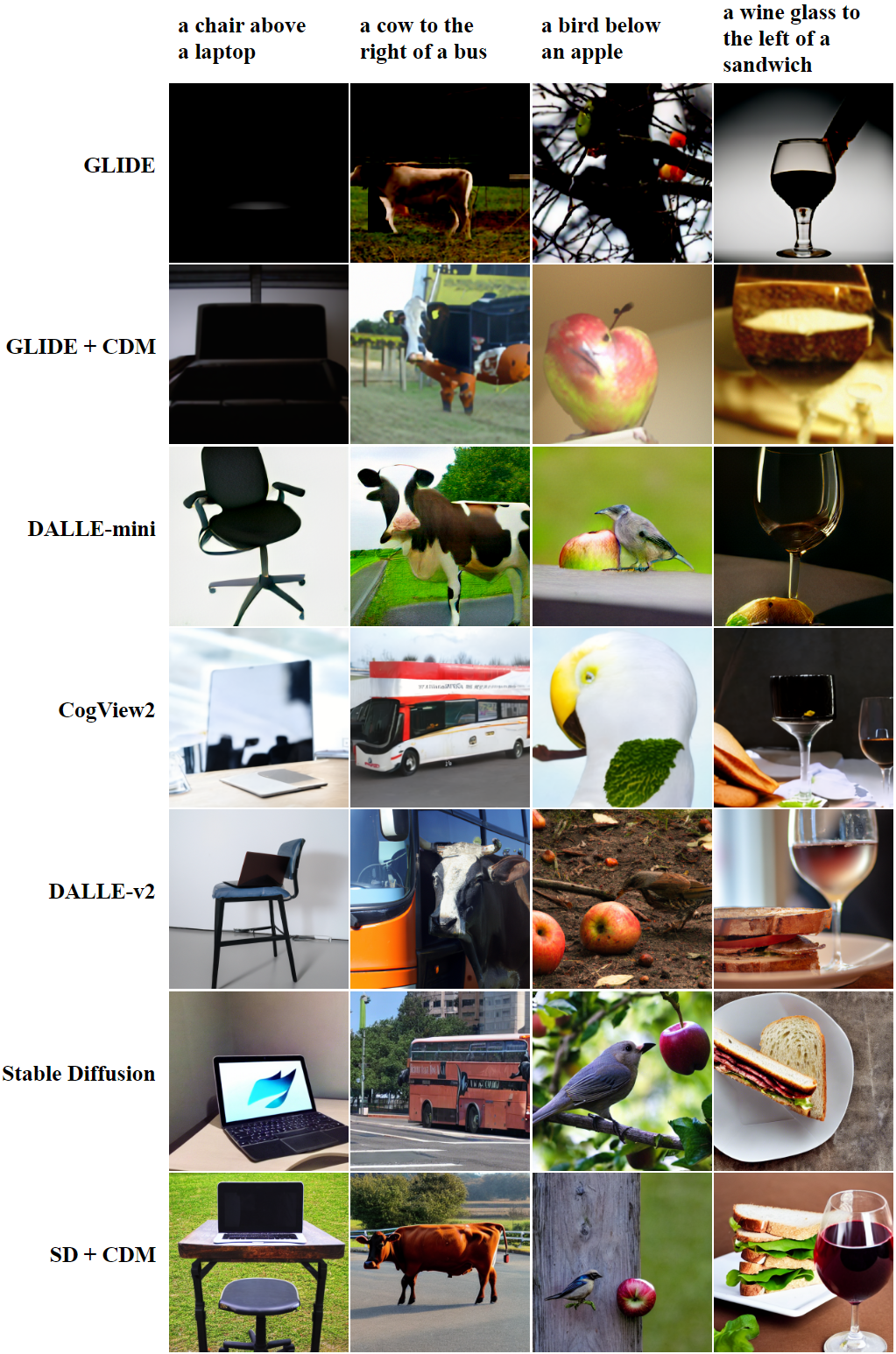}
    \caption{Illustrative examples of text prompts from our SR\textsubscript{2D} dataset and corresponding images generated by each T2I model.}
    \label{fig:all_models_example_1}
\end{figure}

\subsection{Qualitative Results.}
\cref{fig:all_models_example_1} shows examples of images generated by all baselines for each prompt, with more visualizations in the appendix.
Although the photorealism of recent models, such as DALLE-v2, SD, and SD+CDM, is much higher, all models are equally poor at generating accurate spatial relationships.

\subsection{Curious Cases}
\noindent\textbf{Merged Objects.}
\begin{figure}
    \centering
    \includegraphics[width=\linewidth]{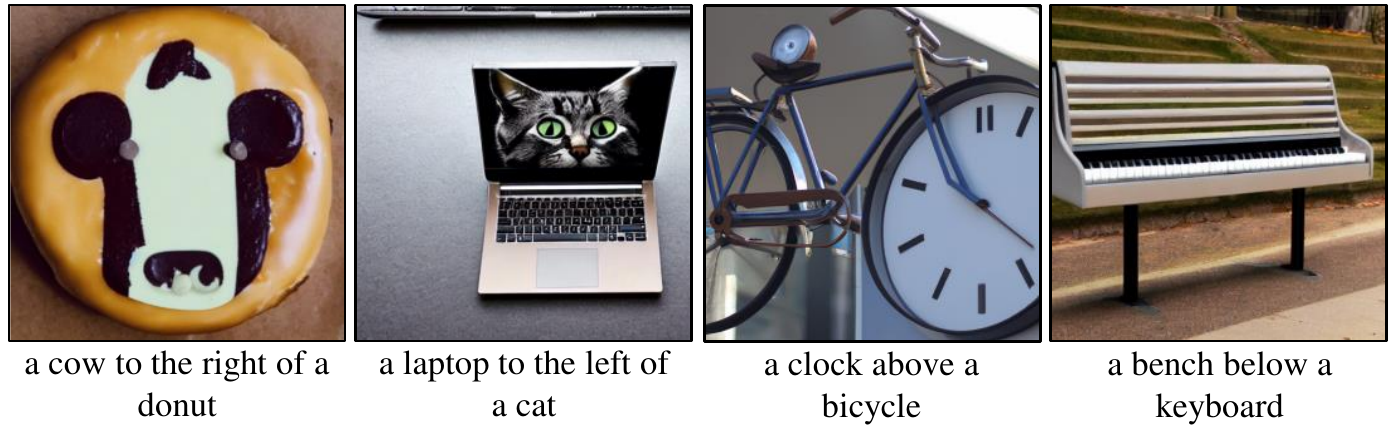}
    \caption{Illustrative examples where the two objects from the text input appear to be merged. From left to right: \textit{a, b, c, d}.}
    \label{fig:merged}
\end{figure}
\cref{fig:merged} shows examples of a few common types of merging between objects that we observed, %in the generated images, 
especially with DALLE-v2.
Common patterns observed include animals being rendered as patterns on inanimate objects (a, b) and both objects retaining their typical shape but getting merged (c, d).
As our human study in \cref{fig:humanstudy_mean} shows, a large proportion (more than 20\%) of images have merged objects -- this poses a significant challenge for generating distinct objects and their relationships using T2I models.

\begin{figure}
    \centering
    \includegraphics[width=\linewidth]{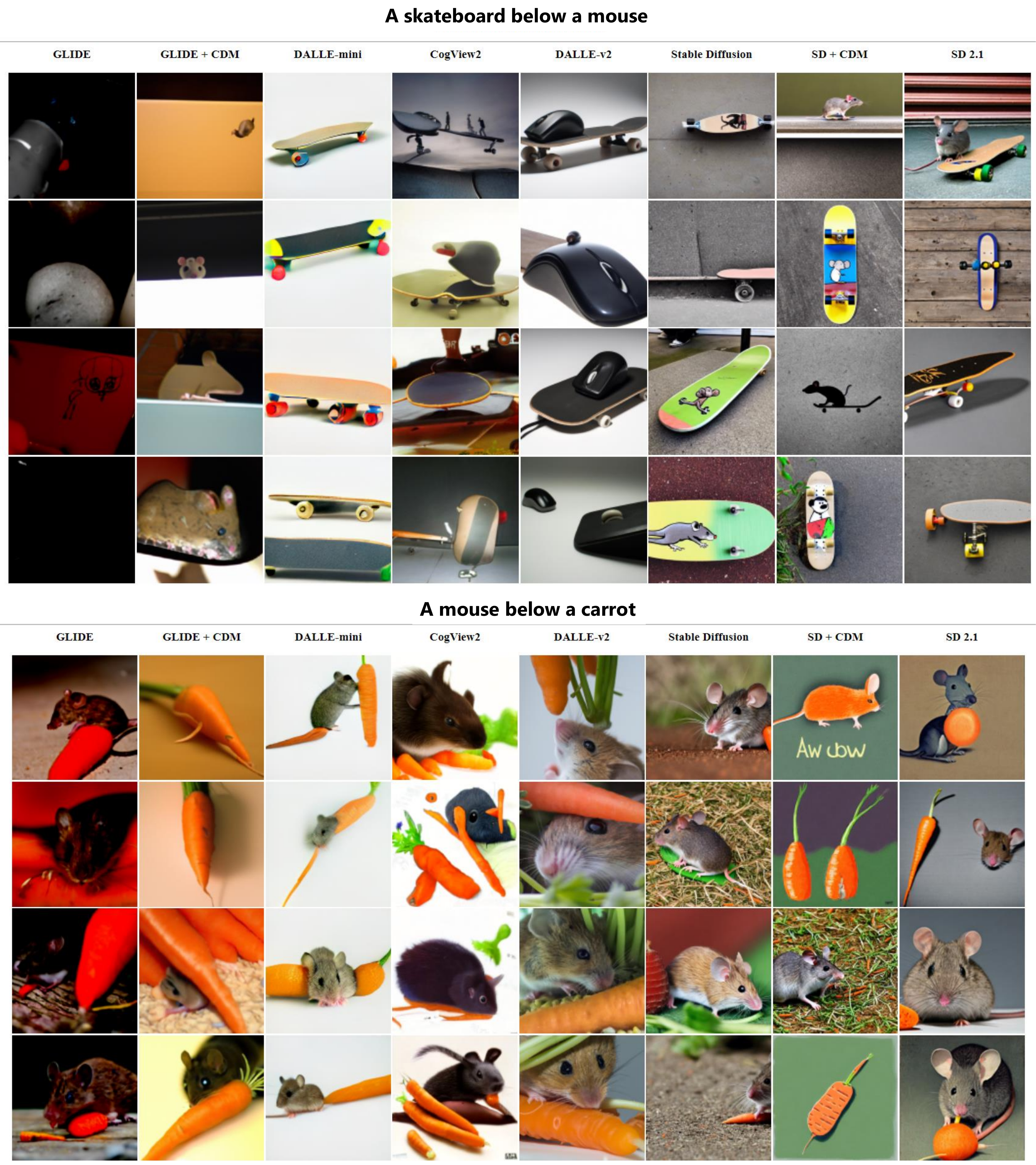}
    \caption{T2I Models can interpret word in different senses under different contexts. Each model seems to prefer a certain word sense in each context.}
    \label{fig:wordsense}
\end{figure}

\medskip\noindent\textbf{Word Sense Ambiguity.}
We also observed curious differences between models when generating images for sentences containing words that have different \textit{senses}, i.e. a different meaning in context. 
For example, the word \textit{``mouse''} is often used to mean the animal mouse and also the computer accessory mouse.
We observed that different models under different contexts interpreted the word as one of the multiple meanings of that word and generated images accordingly.
The first example in Figure \ref{fig:wordsense} shows that for the sentence \textit{``A skateboard below a mouse''}, GLIDE, SD, SD+CDM and SD 2.1 generated the animal mouse whereas CogView2 and DALLE-v2 generated computer mouse.
The SD variants also merged mouse and skateboard in some images by generating a cartoon mouse merged as the design of the skateboard.
However for the second example \textit{``A mouse below a carrot''} all models generated an animal mouse.
Similarly the words \textit{``kite''} and \textit{``keyboard''} also resulted in generation of two senses -- the bird kite vs. the toy kite made of paper and string, and a computer keyboard vs. the musical keys on a piano.
As there are several words in the English language (for instance bass, bow, bat, bank, crane, palm, letter, and so on) where word sense ambiguity could affect the outputs of T2I models, this aspect needs to be studied in depth further to improve the experience of users of T2I models.

\begin{table*}[t]
    \centering
    % \LARGE
    % \resizebox{\linewidth}{!}{
    \begin{tabular}{l rrrr c rrrr}
        \toprule
        \multirow{2}{*}{\textbf{Model}} & \multicolumn{4}{c}{\textbf{VISOR\textsubscript{cond} (\%)}} & \hphantom & \multicolumn{4}{c}{\textbf{Object Accuracy (\%)}}\\
        \cmidrule{2-5} \cmidrule{7-10}
        & left & right & above & below && left & right & above & below \\
        \midrule
        GLIDE       & 57.78 & \textbf{61.71} & 60.32 & 56.24 && 3.10 &     3.46 &      \textbf{3.49} &            3.39\\
        GLIDE + CDM & 65.37 & \textbf{65.46} & 59.40 & 59.84 && \textbf{12.78} &    12.46 &     7.75 &            7.68\\
        DALLE-mini  & 57.89 & 60.16 & \textbf{63.75} & 56.14 && 22.29 &    21.74 &     \textbf{33.62} &          30.74 \\
        CogView2    & \textbf{68.50} & 68.03 & 63.72 & 62.51 && \textbf{20.34} &    19.30 &     17.71 &          16.54\\
        DALLE-v2    & 56.47 & 56.51 & 60.99 & \textbf{63.24} && 64.30 &    64.32 &     \textbf{65.66} &          61.45\\
        SD          & \textbf{64.44} & 62.73 & 61.96 & 62.94 && 29.00 &    29.89 &     \textbf{32.77} &          27.8\\
        SD + CDM    & \textbf{69.05} & 66.52 & 62.51 & 59.94 && 23.66 &    21.17 &     23.66 &          \textbf{24.61}\\
        \bottomrule
    \end{tabular}
    % }
    \caption{Comparison of Visor and OA split by relationship type}
    \label{tab:split_by_rel}
\end{table*}
\begin{figure}
    \centering
    \includegraphics[width=\linewidth]{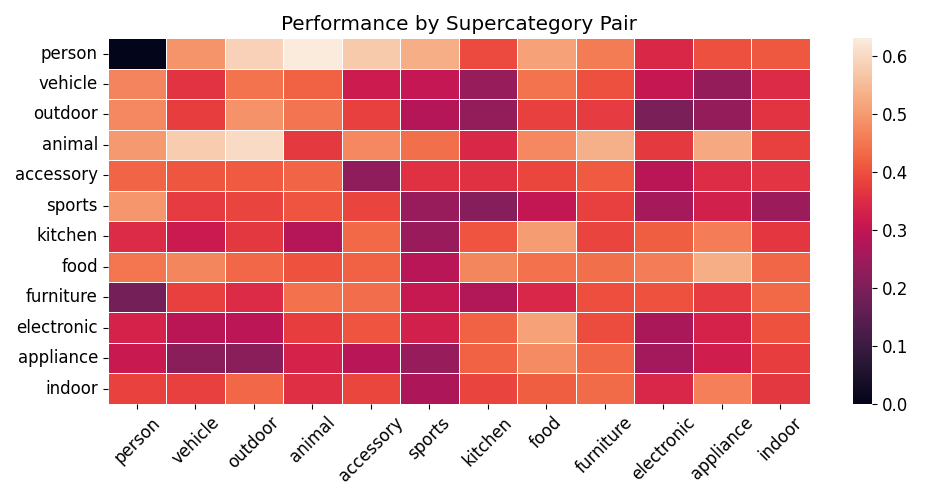}
    \caption{VISOR scores for each supercategory pair.}
    \label{fig:supercategory}
\end{figure}

\subsection{Fine-grained Performance Analysis}
\medskip\noindent\textbf{Performance per relationship} is shown in \Cref{tab:split_by_rel}.
Interestingly, five of the seven models have the best VISOR\textsubscript{cond} scores for horizontal relationships (left or right).
However, five of the seven models have the best object accuracy for vertical relationships (above or below).

\medskip\noindent\textbf{Performance per Supercategory.}  \label{sec:supercategory}
The 80 object categories in SR\textsubscript{2D} belong to 11 MS-COCO ``supercategories''.
We investigate VISOR scores for each supercategory pair and report the results for the best model (DALLE-v2) in \cref{fig:supercategory} (results for other models are in the appendix).
VISOR scores for commonly co-occurring supercategories such as \textit{``animal, outdoor''} are highest whereas unlikely combinations of indoor-outdoor objects such as \textit{``vehicle, appliance''} and \textit{``electronic, outdoor''} have low performance.

\subsection{Correlation with Object Co-occurrence Statistics}
\begin{figure}
    \centering
    \includegraphics[width=\linewidth]{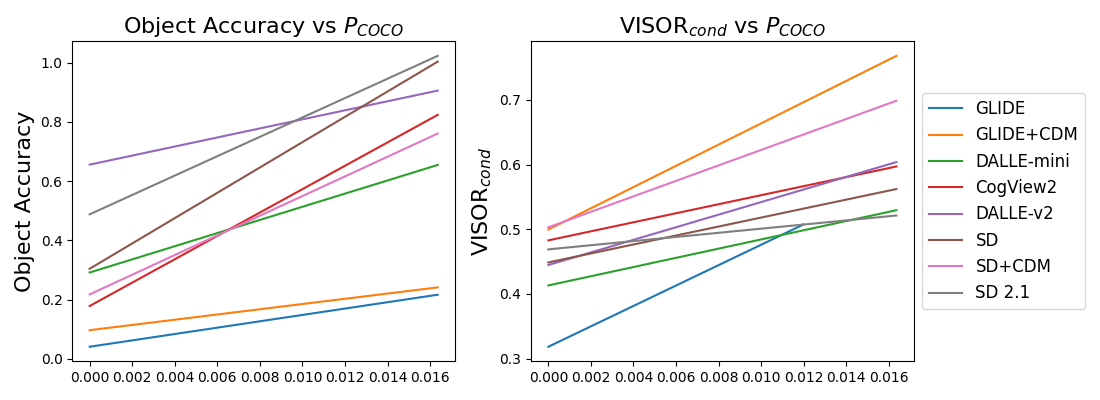}
    \caption{Correlation of our metrics with $P_{COCO}$, the object co-occurrence probability in MS-COCO.}
    \label{fig:coco_correlation}
\end{figure}

The object categories in our dataset span a wide range of commonly occurring objects from MS-COCO such as wild animals, vehicles, appliances, and humans, found in varying contexts, including combinations that do not appear together often in real life. For instance, an elephant is unlikely to be found indoors near a microwave oven.
To understand how object co-occurrence affects VISOR, we first obtain $P_{\textsc{COCO}}(A, B)$, the probability of co-occurrence for each object pair (A, B) as a proxy for real-world object co-occurrence.
Then, we plot the correlation of VISOR and object accuracy for pair (A, B) with its $P_{\textsc{COCO}}(A, B)$.
As \cref{fig:coco_correlation} shows, the correlation is positive for all models, for both OA and VISOR$_{cond}$, implying that the quality of outputs is likely to be better for commonly co-occurring objects, clearly establishing a bias towards real-world likelihood.
This correlation shows the difficulty in generating unlikely relationships such as \textit{``an elephant to the left of a microwave''} even though such unlikely combinations may be desired by creators,
pursuing artistic compositions.

\begin{figure}
    \centering
    \includegraphics[width=1.00\linewidth]{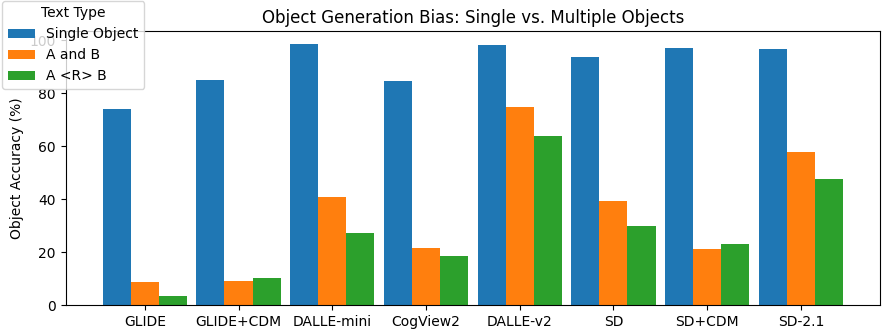}
    \caption{Comparison of object accuracy for text with single and multiple objects reveals a bias towards single objects.}
    \label{fig:objacc_12}
\end{figure}
\begin{figure}
    \centering
    \includegraphics[width=1.00\linewidth]{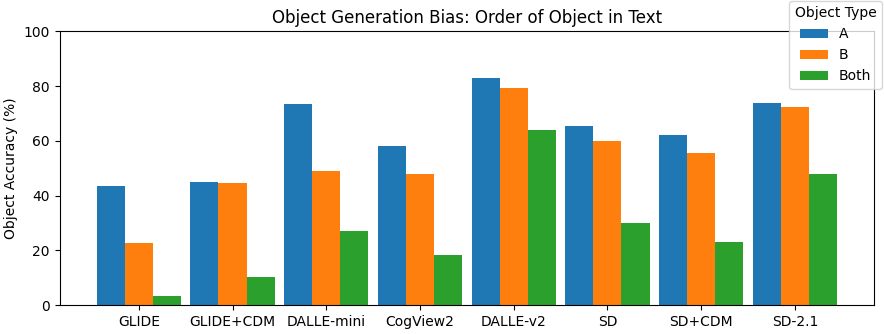}
    \caption{Comparison of object accuracy for object A and B reveals a bias towards A, the first object appearing in the prompt).}
    \label{fig:objacc_AB}
\end{figure}

\subsection{Bias and Consistency}
\medskip\noindent\textbf{Object Generation Bias.}
We compare object accuracy with three types of inputs to generate images: (1) single objects text such as \textit{``an elephant''}, (2) multiple object conjunction such as \textit{``an elephant and a cat''}, and (3) relational texts such as \textit{``an elephant to the right of a cat''}. 
\cref{fig:objacc_12} shows that, for all models, OA is significantly higher for single objects; while composition using conjunction is challenging, systems perform better with this generation than spatial composition.

\medskip\noindent\textbf{Text-Order Bias.}
In \cref{fig:objacc_AB}, we show that for all models, OA for the first mentioned object (A) in the text is significantly higher than OA for the second object (B); generating both objects together is most challenging.

\medskip\noindent\textbf{Consistency between equivalent phrases.}
Ideally, given two equivalent inputs such as \textit{``a cat above a dog''} and \textit{``a dog below a cat''}, the model should generate images with the same spatial relationship.
To evaluate this consistency, we consider cases in which both objects are detected and report the consistency for each relationship type in \Cref{tab:consistency}.
Surprisingly, the best model \textit{DALLE-v2 is the least consistent}, while CogView2 is the most consistent model.
This result shows that merely rephrasing the input can have a large influence on the spatial correctness of the output.

\begin{table}[t]
    \centering
    \footnotesize
    \begin{tabular}{@{}l ccccc@{}}
        \toprule
        {\textbf{Model}} 
         & left & right & above & below & Average \\ 
        \midrule
        GLIDE       & \underline{45.90} & 58.93 & 63.16 & 52.63 & 55.16 \\
        GLIDE + CDM & 61.99 & 59.15 & 54.79 & 56.15 & 58.02 \\
        DALLE-mini  & 54.75 & 52.28 & 54.64 & 55.77 & 54.36 \\
        CogView2    & \textbf{67.32} & 65.38 & \textbf{65.67} & \textbf{66.95} & \textbf{66.33} \\
        DALLE-v2    & 48.81 & \underline{48.10} & \underline{48.72} & \underline{48.15} & \underline{48.45} \\
        SD & 58.71  & 61.36 & 55.36 & 55.39 & 57.71 \\
        SD + CDM    & 64.69 & \textbf{65.71} & 61.35 & 57.71 & 62.37 \\ 
        SD 2.1 & 53.96 & 55.50 & 54.73 & 54.38 & 54.64\\
        \bottomrule
    \end{tabular}
    % \vspace{1pt}
    \caption{
        Consistency (\%) of generated spatial relationships for equivalent inputs.
        \textbf{Bold}: highest, \underline{Underline}: lowest consistency.
    }
    \label{tab:consistency}
\end{table}

\subsection{Effect of Linguistic Variation}
\begin{figure}
    \centering
    \includegraphics[width=\linewidth]{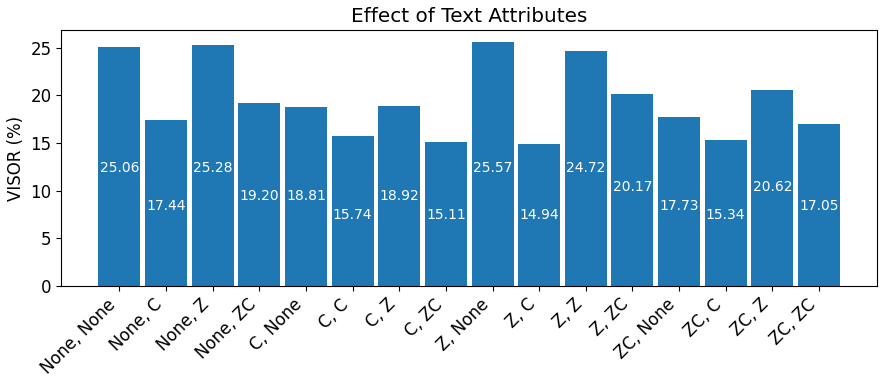}
    \caption{Comparing VISOR performance with different combinations of attributes. \textit{``Z, ZC''} indicates a prompt describing object A with a size attribute and object B with both size and color.}
    \label{fig:effect_of_attributes}
\end{figure}

\medskip\noindent\textbf{Effect of Attributes on Spatial Understanding.}
We conduct a case study with Stable Diffusion (SD) to seek an understanding of the impact of sentence complexity on a model's VISOR performance.
We increase the complexity of text prompts by randomly assigning two attributes (size \texttt{Z} and color \texttt{C}) to the object category, via templates of the form 
\texttt{[Z\textsubscript{A}][C\textsubscript{A}]<A> <R> [Z\textsubscript{B}][C\textsubscript{B}]<B>}.
We focus on 11 object categories representative of each supercategory in COCO, 8 colors, and 4 sizes. 
As shown in \cref{fig:effect_of_attributes}, compared to generation without attributes, there is a drop in performance in 13 out of 15 types of attribute combinations.
Addition of the color attribute (C) leads to a large drop in performance. Adding size descriptors (Z) may improve performance.
While concurrent work~\cite{feng2022training} has reported difficulty in attribute binding, our analysis suggests that attributes may negatively influence spatial compositionality.

\begin{table}[t]
    \centering
    \Large
    \resizebox{\linewidth}{!}{
    \begin{tabular}{@{}l rr r cccc@{}}
        \toprule
     & \multirow{2}{*}{\textbf{OA (\%)}} & \multicolumn{6}{c}{\textbf{VISOR (\%)}} \\
         \cmidrule{3-8}
        \textbf{Prompt Type} & & \textbf{uncond} & \textbf{{cond}} & \textbf{1} & \textbf{2} & \textbf{3} & \textbf{4}\\
        \midrule
        Phrases & 29.86 & 18.81 & 62.98 & 46.60 & 20.11 &  6.89 & 1.63 \\
        Sentences       & 32.48 & 20.67 & 63.64 & 48.54 & 22.94 & 8.92 & 2.25\\
        Split Sentences & 24.98 & 16.44 & 65.82 & 41.91 & 16.29 & 5.66 & 1.91 \\ 
        \bottomrule
    \end{tabular}
    }
    \caption{Effect of prompt variations on OA and VISOR scores. All three versions use the same Stable Diffusion (SD) model .}
    \label{tab:effect_prompt_variations}
\end{table}
\medskip\noindent\textbf{Effect of Rephrased Text Prompts.}
We compare four variations of the prompt: 
\begin{enumerate}
\item \textit{phrases}: the default version of SR$_{2D}$ used in \cref{tab:visor_main} (\eg ``a cat to the left of a dog''); 
\item \textit{sentences} (\eg ``There is a cat to the left of a dog''), and 
\item \textit{split sentences} (\eg ``There is a cat to the left.\ There is a dog to the right'').
Compared to phrases, 
\item \textit{rephrasing using Language Models}: variations of our prompts from {GPT3.5-Turbo}
\end{enumerate}

\cref{tab:effect_prompt_variations} shows that compared to phrases, sentences have higher OA and VISOR\textsubscript{cond}; while split sentences have lower OA and higher VISOR\textsubscript{{cond}}.
Table \ref{tab:results_gpt_rephrased} shows the results of a small-scale experiment where  we obtained 3 variations from {GPT3.5-Turbo} for 500 SR\textsubscript{2D} prompts (examples shown below) and generated images using SD 2.1.
For instance the prompt \textit{``an apple below a skateboard''} was rephrased as \textit{``beneath the skateboard lies an apple''} or \textit{``a cat to the right of a toaster''} was rephrased as \textit{``on the right side of a toaster, there is a cat''}.
The results suggest that with GPT-rephrased prompts, OA is higher but VISOR is lower -- hinting that prompt engineering with language models might help object-level evaluation, but still may not enhance spatial understanding.

\begin{table}[t]
    \centering
    \Large
    \resizebox{\linewidth}{!}{
    \begin{tabular}{@{}l rr r cccc@{}}
        \toprule
         & \multirow{2}{*}{\textbf{OA (\%)}} & \multicolumn{6}{c}{\textbf{VISOR (\%)}} \\
         \cmidrule{3-8}
        \textbf{Dataset} & & \textbf{uncond} & \textbf{{cond}} & \textbf{1} & \textbf{2} & \textbf{3} & \textbf{4}\\
        \midrule
        Original       & 45.83 & {30.47} & {66.49} & 65.69 & 35.62 & 16.34 & 4.25\\
        GPT-rephrased  & {46.16} & 29.99 & 64.98 & 63.54 & 35.10 & 16.59 & 4.74\\
        \bottomrule
    \end{tabular}
    }
    \caption{Caption}
    \label{tab:results_gpt_rephrased}
\end{table}
These observations of differences in performance by simply rephrasing the prompt are a signal that prompt engineering for grounded generation such as spatial aspects is a promising future direction to be investigated.

\section{Discussion and Conclusion}\label{sec:discussion}
We studied the spatial capabilities of text-to-image generators by introducing spatial relationship metrics (VISOR measures), building a dataset (SR\textsubscript{2D}), and developing an automated evaluation pipeline.
Our experiments reveal that existing T2I models have poor spatial interpretation and rendering abilities, as characterized by their low VISOR scores, making them unreliable for uses that depend on the correctness in generated images of spatial relations specified in prompts.
Our analysis also reveals several biases and artifacts of T2I models, such as proclivity for generating single objects (especially the first mentioned object), correlation of spatial correctness with likelihood of object co-occurrence, sensitivity to equivalent phrasings of spatial relations, and negative influences of the inclusion in prompts of several commonly used modifiers.  We hope that the metrics, methods, and dataset will help to stimulate a stream of research on the spatial rendering capabilities of generative models, leading to enhancements of these capabilities over time. For uses of today's technologies, we hope our findings can provide creators with guidance for prompt engineering.
We note that the SR\textsubscript{2D} data generation pipeline can be extended to study spatial relationships of more than two objects, including three-dimensional and complex relations such as \textit{inside, outside, contains, behind, in front, covers, touching}, as well as semantic and action-based relationships.

One of major findings of this paper is the big gap in expectations from generative models and the grounded abilities of those models.
At the same time, we also note that evaluation itself is a significant challenge and our paper is an effort towards revealing those challenges and creating a path towards holistic evaluation.
While general purpose scores are easier to implement, they can't fully assess each skill or ability that is of interest to practitioners.
Our dataset and evaluation pipeline offers users a focused evaluation of spatial abilities.
We hope the VISOR metric will serve as a complement to prior metrics that evaluate photorealism and image-text similarity.
We envision T2I evaluation to proceed in the direction of multi-faceted evaluations that are prevalent in the natural language processing community, such as the GLUE  \cite{wang2018glue} and Super-GLUE \cite{wang2019superglue} benchmarks for natural language understanding.

\bibliographystyle{IEEEtran}
\bibliography{tgokhale}

\appendices

In this appendix we provide more details about the \sr\ dataset, additional experimental results and analyses, updates to the benchmark, proofs, and qualitative examples.

\section{Additional Details about the SR\textsubscript{2D} Dataset}
\textbf{List of COCO categories.} 
In the \sr\ dataset we use 80 object categories from the MS-COCO dataset as the set of objects $\mathcal{C}$.
The box below lists all of these categories.

\smallskip
\setlength{\fboxsep}{0pt}
\noindent\fbox{%
    \parbox{\linewidth}{
        \centering
        \smallskip
        \small
        \tt
        person, bicycle, car, motorcycle, airplane, bus,
	    train, truck, boat, traffic light, fire hydrant,
	    stop sign, parking meter, bench, bird, cat, dog, horse,
	    sheep, cow, elephant, bear, zebra, giraffe, backpack,
	    umbrella, handbag, tie, suitcase, frisbee, skis,
	    snowboard, sports ball, kite, baseball bat, baseball glove,
	    skateboard, surfboard, tennis racket, bottle, wine glass,
	    cup, fork, knife, spoon, bowl, banana, apple, sandwich,
	    orange, broccoli, carrot, hot dog, pizza, donut, cake,
	    chair, couch, potted plant, bed, dining table,
	    toilet, tv, laptop, mouse, remote, keyboard,
	    cell phone, microwave, oven, toaster, sink, refrigerator,
	    book, clock, vase, scissors, teddy bear, hair drier,
	    toothbrush
        \smallskip
    }
}

\medskip
\noindent\textbf{List of COCO supercategories.} 
In Figure \ref{fig:supercategory} in the main paper, we presented results for each supercategory. The box below lists these eleven supercategories.

\smallskip
\setlength{\fboxsep}{0pt}
\noindent\fbox{%
    \parbox{\linewidth}{
        \centering
        \smallskip
        \small
        \tt
        person, vehicle, outdoor, animal, accessory, sports, kitchen, food, furniture, electronic, appliance, indoor
        \smallskip
    }
}

\section{Experimental Setup for ``Effect of Attributes on Spatial Understanding''.} 
In Figure \ref{fig:effect_of_attributes}, we compared VISOR scores for text prompts with and without size and color attributes.
We used one object category for 11 supercategories for this analysis.
Note that we ignore the \texttt{person} category since colors are not typically used as attributes for people (for example ``purple person'') and to avoid any potentially racist stereotypes associated with skin color to percolate into our generated images.
Table \ref{tab:supercat_examples} shows the object categories that we used:
\begin{table*}[!t]
    \centering
    \resizebox{\linewidth}{!}{
    \begin{tabular}{@{}|c|c|c|c|c|c|c|c|c|c|c|c|@{}}
        \hline
        \textbf{Supercategory} & vehicle & outdoor & animal & accessory & sports & kitchen & food & furniture & electronic & appliance & indoor \\
        \hline
        \textbf{Category} & car & bench & dog & suitcase & sports ball & cup & cake & chair & laptop & microwave & book\\
        \hline
    \end{tabular}
    }
    \caption{Object categories that we used for each supercategory for analyzing the effrt of attributes on spatial understansing.}
    \label{tab:supercat_examples}
\end{table*}

\section{Proof of Equation \ref{eq:visor_proof}}
\noindent Equation \ref{eq:visor_proof} states the following relationship between \textsc{VISOR} and \textsc{VISOR}\textsubscript{n}.
\begin{equation}
    \small
    \textsc{VISOR} = \frac{1}{N} \sum_{n=0}^{N} n (\textsc{VISOR}_n - \textsc{VISOR}_{n+1}).
\end{equation}
\noindent\textbf{Proof.}
First, we restate the definitions of \textsc{VISOR} and \textsc{VISOR}\textsubscript{n} below.
\begin{equation}
    \small
    \textsc{VISOR} (x, A, B, R) = 
    \begin{cases}
    1, \quad \text{if } R_{gen}=R \cap A \cap B\\
    0, \quad \text{otherwise}.
    \end{cases}
    \label{eq:supp_VISOR_uncond}
\end{equation}

\begin{equation}
    \small
    \textsc{VISOR}_n (x,A,B,R) =
    \begin{cases}
        1,~~ \text{if }\sum\limits_{i=1}^{N}\textsc{VISOR}(x_i, A, B,R) \geq n \\
        0,~~ \text{otherwise}.
    \end{cases}
    \label{eq:supp_VISOR_n} 
\end{equation}

Let $T$ be the total number of text prompts used for evaluating VISOR of a text-to-image model.
For a model that generates $N$ images per prompt, we have $NT$ total generated images.
Let $V$ be the number of images for which $\textsc{VISOR}=1$, i.e. images for which $R=R_{gen}\cap A \cap B$.
From \cref{eq:supp_VISOR_uncond}, it is clear that 
\begin{equation}\label{eq:supp_visor_frac}
    \textsc{VISOR} = \frac{V}{NT}.
\end{equation}

\noindent Let $P_n$ be the number of prompts for which \textit{at least} $n$ generated images were spatially correct.
From \cref{eq:supp_VISOR_n} we can say:
$$\textsc{VISOR}_n = \frac{P_n}{T}$$
$\Rightarrow$ $P_n - P_{n+1}$ is the number of prompts for which \textit{exactly} $n$ images are correct.\\
$\Rightarrow$ $\sum_{i=0}^{N} n(P_n - P_{n+1})$ is the number of generated images which are spatially correct.\\
\begin{align}
\Rightarrow V &= \sum_{i=0}^{N} n(P_n - P_{n+1})\nonumber\\
\Rightarrow \frac{V}{NT}     &= \frac{1}{NT}\sum_{i=0}^{N} n(P_n - P_{n+1})\nonumber\\
\Rightarrow \textsc{VISOR}   &= \frac{1}{N}\sum_{i=0}^{N} n(\textsc{VISOR}_n - \textsc{VISOR}_{n+1}) \nonumber\\
& \qquad{\textit{from \cref{eq:supp_visor_frac,eq:supp_VISOR_n}}}\nonumber
\end{align}

\section{Additional Experiments}
\subsection{Performance Per Supercategory.}
\begin{figure*}[t]
    \centering
    \begin{subfigure}[t]{0.49\linewidth}\includegraphics[width=\linewidth]{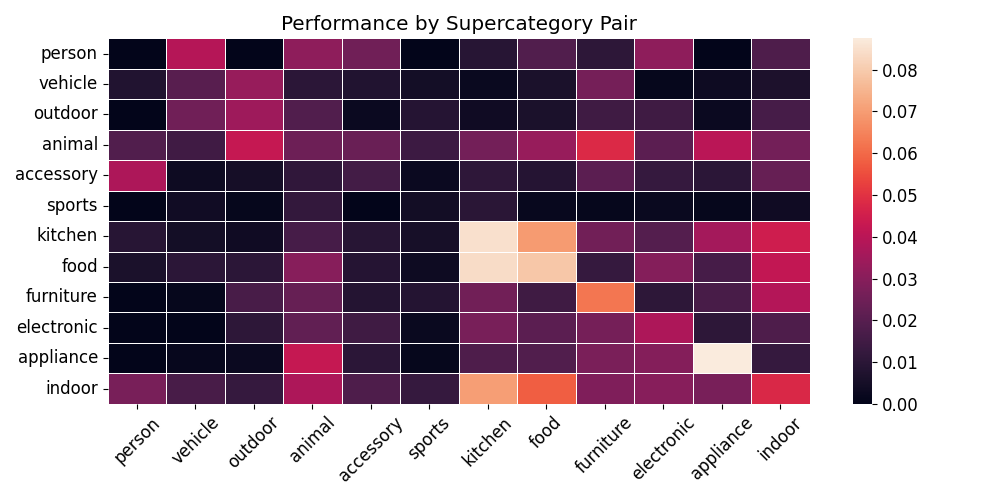}
    \caption{GLIDE}
    \end{subfigure} 
    \begin{subfigure}[t]{0.49\linewidth}\includegraphics[width=\linewidth]{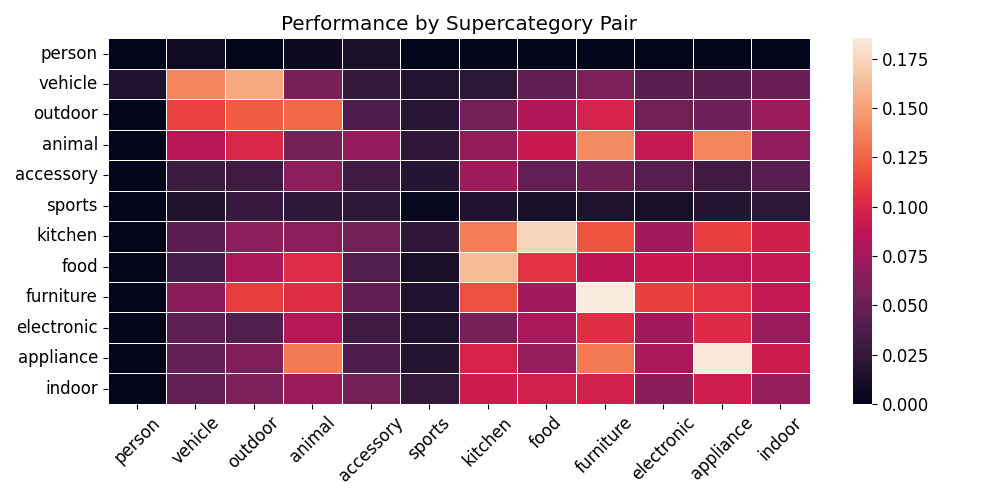}
    \caption{GLIDE+CDM}
    \end{subfigure}
    \begin{subfigure}[t]{0.49\linewidth}\includegraphics[width=\linewidth]{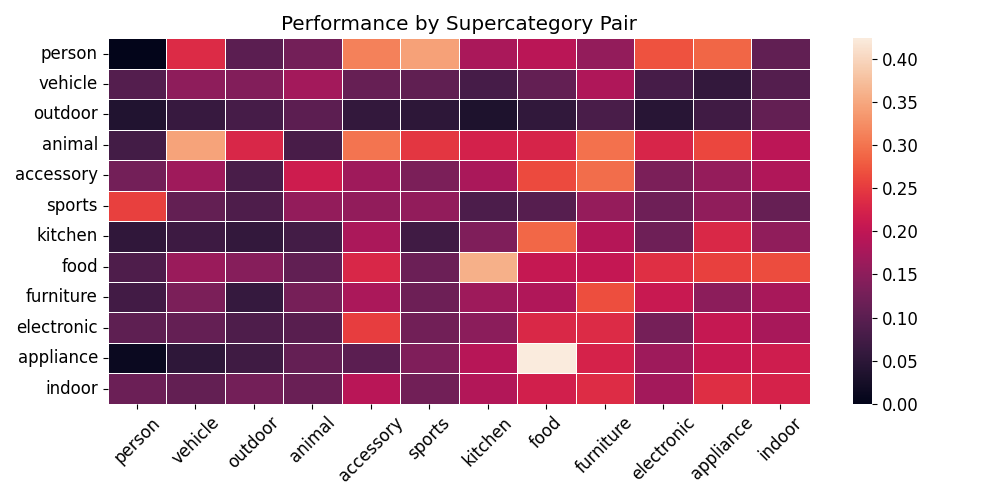}
    \caption{DALLE-mini}
    \end{subfigure}
    \begin{subfigure}[t]{0.49\linewidth}\includegraphics[width=\linewidth]{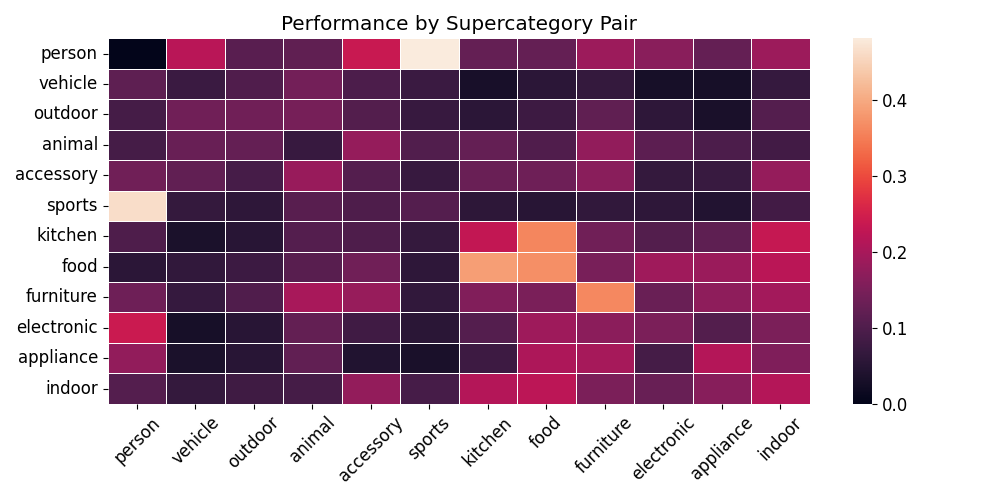}
    \caption{CogView2}
    \end{subfigure}
    \begin{subfigure}[t]{0.49\linewidth}\includegraphics[width=\linewidth]{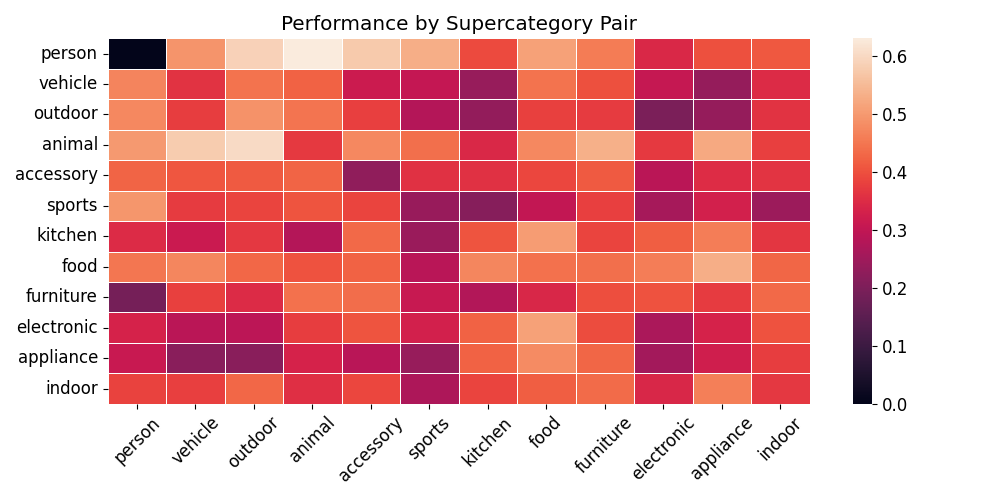}
    \caption{DALLE-v2}
    \end{subfigure}
    \begin{subfigure}[t]{0.49\linewidth}\includegraphics[width=\linewidth]{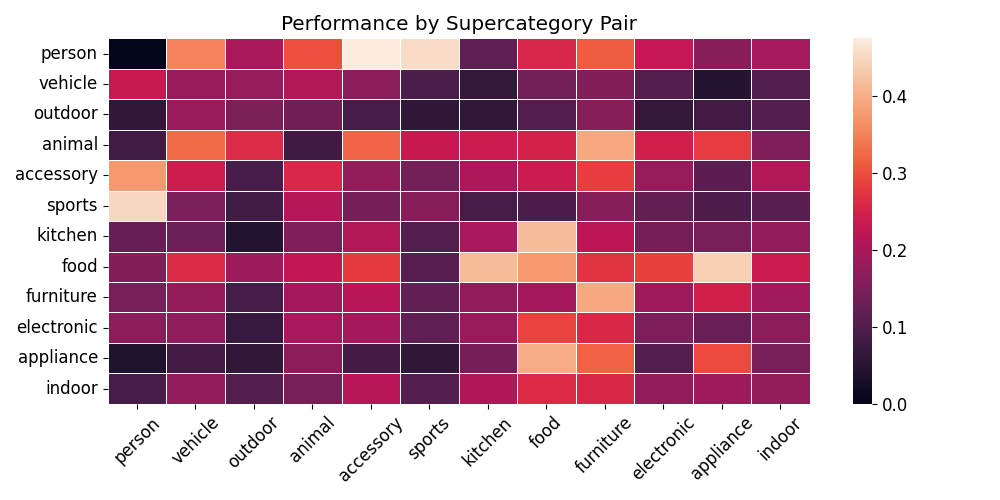}
    \caption{Stable Diffusion}
    \end{subfigure}
    \begin{subfigure}[t]{0.49\linewidth}\includegraphics[width=\linewidth]{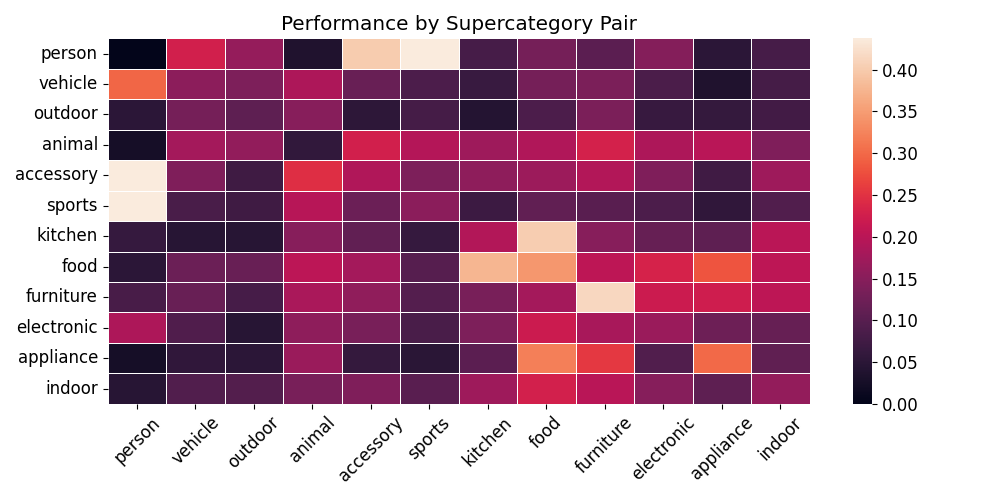}
    \caption{SD+CDM}
    \end{subfigure}
    \caption{VISOR scores of each model split by supercategory pairs.}
    \label{fig:supp_supercategory}
\end{figure*}

\cref{fig:supp_supercategory} shows the performance per supercategory for all seven models.

\subsection{Benchmarking using COCO-finetuned Object Detectors.}

\begin{table}[t]
    \centering
    \resizebox{\linewidth}{!}{
    \begin{tabular}{@{}l rr r rrrr@{}}
        \toprule
         & \multirow{2}{*}{\textbf{OA (\%)}} & \multicolumn{6}{c}{\textbf{VISOR (\%)}} \\
         \cmidrule{3-8}
        \textbf{Model} & & \textbf{uncond} & \textbf{{cond}} & \textbf{1} & \textbf{2} & \textbf{3} & \textbf{4}\\
        \midrule
        GLIDE       &  0.23 & 2.54 &  0.12 &  0.47 & 0.02 & 0.00 & 0.00 \\
        GLIDE+CDM   &  1.49 & 5.09 &  0.82 &  2.90 & 0.33 & 0.04 & 0.01 \\
        DALLE-mini  &  6.91 & 3.16 &  3.67 & 11.34 & 2.65 & 0.62 & 0.08 \\
        CogView2    &  4.75 & 6.86 &  2.70 &  8.85 & 1.62 & 0.29 & 0.04 \\
        DALLE-v2    & 14.80 & 3.92 &  7.98 & 22.84 & 6.95 & 1.80 & 0.32 \\
        SD          & 14.17 & 7.14 &  8.09 & 23.37 & 6.91 & 1.73 & 0.37 \\
        SD+CDM      & 11.06 & 9.34 &  6.56 & 19.49 & 5.13 & 1.40 & 0.22 \\
        \bottomrule
    \end{tabular}
    }
    \caption{Benchmarking performance of all models in terms of object accuracy (OA) and each version of VISOR, with DETR-ResNet-50 (trained on MS-COCO) as the object detector}
    \label{tab:visor_detr50}
\end{table}

In the main paper, we used the open-vocabulary object detector OWL-ViT \cite{minderer2022simple} as the oracle to localize objects and identify their spatial relationships.
In this section, we replicate these benchmarking experiments by using DETR-Resnet-50~\cite{carion2020end}, finetuned on the MS-COCO dataset.
Results are shown in \cref{tab:visor_detr50}.
The findings are similar using OWLViT.
In terms of OA and VISOR (unconditional), DALLE-v2~\cite{ramesh2022hierarchical} is the best performing model.
However conditional VISOR for DALLE-v2 is low, but higher for CogView~\cite{ding2022cogview2}, SD~\cite{rombach2022high}, and SD+CDM~\cite{liu2022compositional}.
This result shows that irrespective of the object detector, the relative performance comparison and rankings of models obtained via VISOR are consistent.
While the oracle object detector does set an upper-bound for VISOR, it can be seamlessly swapped with any newer and more sophisticated object detectors that may be developed in the future.

\subsection{Effect of Confidence Threshold.}
\begin{figure}[t]
    \centering
    \includegraphics[width=\linewidth]{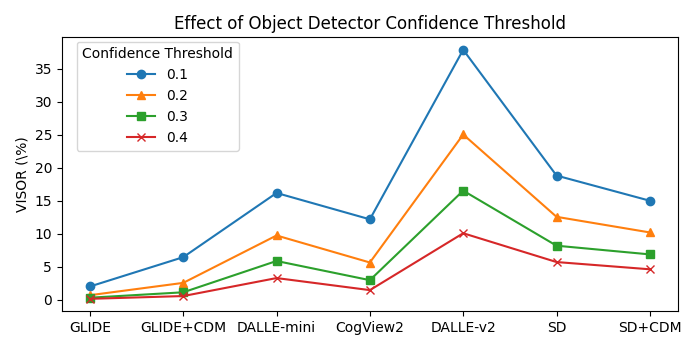}
    \caption{Effect of the confidence threshold of the oracle object detector. The relative ranking of models is consistent irrespective of the confidence threshold.}
    \label{fig:effect_threshold}
\end{figure}

We study how the confidence threshold of the object detector affects VISOR performance. 
In \cref{fig:effect_threshold} we plot the VISOR scores for each model for four values of confidence threshold: $0.1$, $0.2$, $0.3$, and $0.4$.
For a higher (stricter) threshold, naturally, the VISOR score is lower, since fewer objects will be detected than at a lower confidence value.
However, the trend and relative performance of the models are identical irrespective of the confidence threshold, leading to consistency in the rankings of models in terms of VISOR score, validating the use of oracle object detectors for computing VISOR score.
This also implies that in the future, more sophisticated object detection models that may be developed can be incorporated in the VISOR computation pipeline to replace older detectors.

\begin{table*}[t]
    \centering
    \scriptsize
    \resizebox{\linewidth}{!}{
    \begin{tabular}{@{}ll@{}}
        \hline
        \textbf{Original} & \textbf{GPT-Variation} \\
        \hline
        an apple below a skateboard
            & (1) Beneath the skateboard lies an apple. \newline 
              (2) An apple rests underneath a skateboard.\\
        \hline 
        a cat to the right of a toaster 
            & (1) A cat positioned on the right side of a toaster. \newline
              (2) On the right side of a toaster, there is a cat.\\
        \hline
    \end{tabular}
    }
    \caption{GPT-rephrased versions of SR\textsubscript{2D} prompts.}
    \label{tab:supp_sr2d_gpt}
\end{table*}

\subsection{\texttt{UPDATE}: Comparison with new spatially-focused methods.}
After the first version of this preprint was released, more variants of text-to-image generation models have been developed.
Two of these: Structured Diffusion \cite{feng2022training} and Attend-and-Excite \cite{chefer2023attendandexcite} are focused on spatial aspects of the image and are therefore very relevant to the benchmark.
The availability of our dataset and metric will aid in comparing these or future models in terms of spatial reasoning.
Table \ref{tab:supp_new_models} shows the comparitive results with the models in our previous study.
\begin{table}[t]
    \centering
    \LARGE
    \resizebox{\linewidth}{!}{
    \begin{tabular}{@{}l rr r cccc@{}}
        \toprule
         & \multirow{2}{*}{\textbf{OA (\%)}} & \multicolumn{6}{c}{\textbf{VISOR (\%)}} \\
         \cmidrule{3-8}
        \textbf{Model} & & \textbf{uncond} & \textbf{{cond}} & \textbf{1} & \textbf{2} & \textbf{3} & \textbf{4}\\
        \midrule
        GLIDE \cite{nichol2021glide}           &  3.36 &  1.98 & 59.06 &  6.72 & 1.02 & 0.17 & 0.03\\ 
        GLIDE + CDM \cite{liu2022compositional}    & 10.17 &  6.43 & 63.21 & 20.07 &  4.69 &  0.83 & 0.11 \\
        DALLE-mini \cite{dayma2021dallemini}    & 27.10 & 16.17 & 59.67 & 38.31 & 17.50 &  6.89 & 1.96 \\
        CogView2 \cite{ding2022cogview2}       & 18.47 & 12.17 & \textbf{65.89} & 33.47 & 11.43 &  3.22 & 0.57 \\
        DALLE-v2 \cite{ramesh2022hierarchical}       & \textbf{63.93} & \textbf{37.89} & 59.27 & \textbf{73.59} & \textbf{47.23} & \textbf{23.26} & \textbf{7.49} \\
        SD \cite{rombach2022high}             & 29.86 & 18.81 & 62.98 & 46.60 & 20.11 &  6.89 & 1.63 \\
        SD + CDM \cite{liu2022compositional}      & 23.27 & 14.99 & 64.41 & 39.44 & 14.56 &  4.84 & 1.12\\
        SD 2.1      & 47.83 & 30.25 & 63.24 & 64.42 & 35.74 & 16.13 &  4.70 \\  
        Structured Diffusion \cite{feng2022training}    & 28.65 & 17.87 & 62.36 & 44.70 & 18.73 & 6.57 & 1.46 \\
        Attend-and-Excite \cite{chefer2023attendandexcite}      & 42.07 & 25.75 & 61.21 & 49.29 & 19.33 & 4.56 & 0.08\\
        \bottomrule
    \end{tabular}
    }
    \caption{Performance of new spatially-focused models on our benchmark.}
    \label{tab:supp_new_models}
\end{table}

\begin{table}
    \centering
    \resizebox{\linewidth}{!}{
    \begin{tabular}{@{}l rr r cccc@{}}
        \toprule
         & \multirow{2}{*}{\textbf{OA (\%)}} & \multicolumn{6}{c}{\textbf{VISOR (\%)}} \\
         \cmidrule{3-8}
        \textbf{Dataset} & & \textbf{uncond} & \textbf{{cond}} & \textbf{1} & \textbf{2} & \textbf{3} & \textbf{4}\\
        \midrule
        Original       & 45.83 & {30.47} & {66.49} & 65.69 & 35.62 & 16.34 & 4.25\\
        GPT-rephrased  & {46.16} & 29.99 & 64.98 & 63.54 & 35.10 & 16.59 & 4.74\\
        \bottomrule
    \end{tabular}
    }
    \caption{Comparison between performance of SD2.1.\ on the original SR\textsubscript{2D} dataset and its GPT-rephrased versions.}
    \label{tab:supp_results_gpt}
\end{table}

\subsection{Linguistic Variations with Large Language Models.}
Our SR\textsubscript{2D} dataset offers a controlled evaluation without ambiguity or noise.
However recent advances in generative language models have given us access to automated ways to rephrase our prompts.
In a small experiment, we obtaining 3 variations from {GPT3.5-Turbo} for 500 SR\textsubscript{2D} prompts and generated images using Stable Diffusion 2.1.
Examples of the GPT-rephrased text promps are shown in \cref{tab:supp_sr2d_gpt}.
We compare VISOR scores of GPT-rephrased vs.\ original prompts in \cref{tab:supp_results_gpt} and find that with GPT-rephrased prompts, OA is higher but VISOR is lower -- hinting that prompt engineering with LLMs might help object-level evaluation, but still may not enhance spatial understanding.

\section{Survey of Prior Work on T2I Evaluation}
\begin{figure}[!h]
    \centering
    \includegraphics[width=\linewidth]{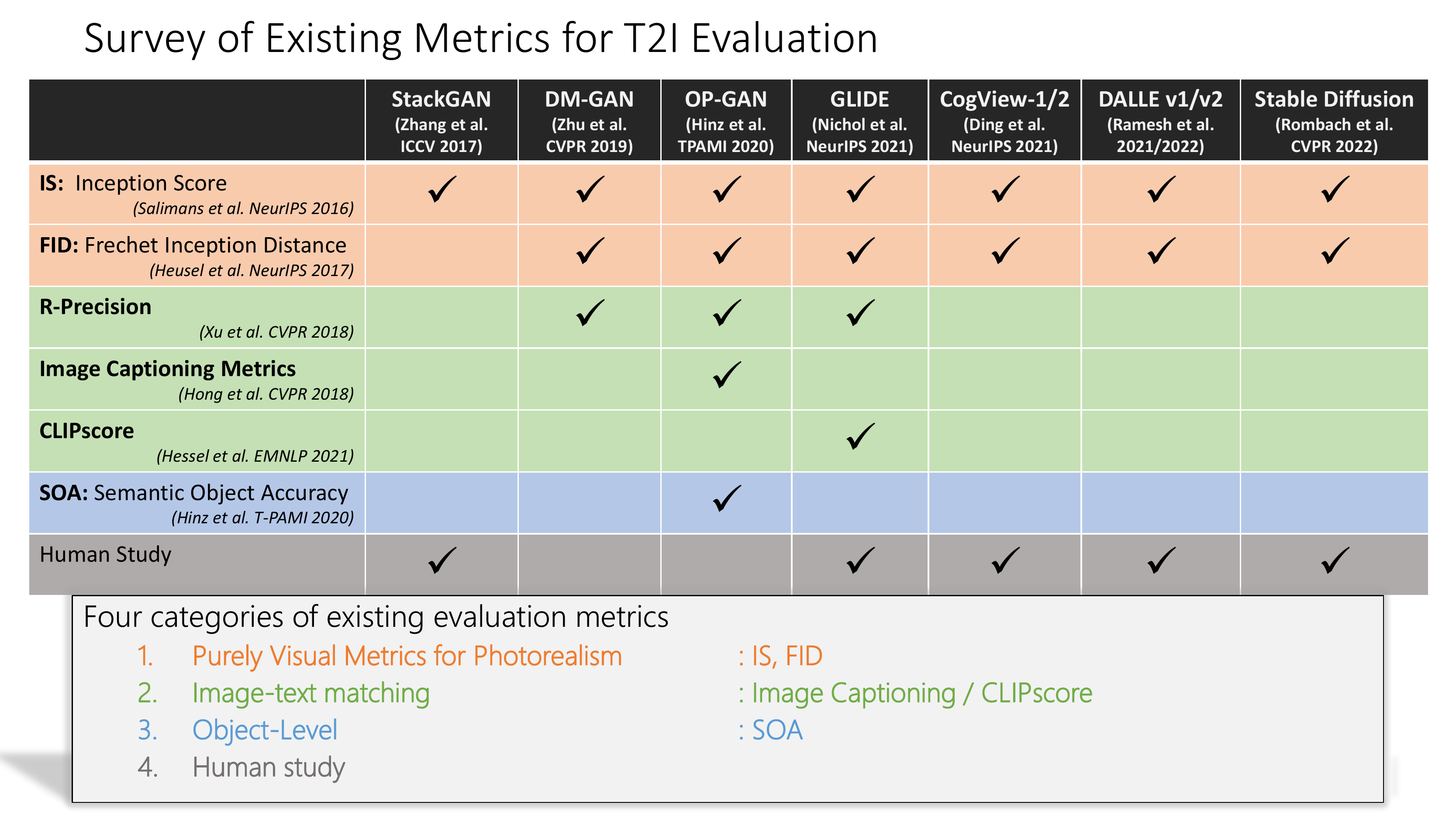}
    \caption{Overview of previous T2I evaluation metrics.}
    \label{fig:supp_metrics_overview}
\end{figure}

Text-to-image synthesis is a relatively new area of research but has seen an explosion in interest and unprecedented improvements in the quality of generated outputs.
In \cref{fig:supp_metrics_overview} we overview existing evaluation metrics for T2I and their used by seminal T2I models from 2017 to 2022.
We categorize these metrics into four broad categories: purely visual metrics, image-text matching, object-level evaluation, and human studies.

\section{Visualization of Generated Images}
\begin{figure*}
    \centering
    \includegraphics[width=0.93\linewidth]{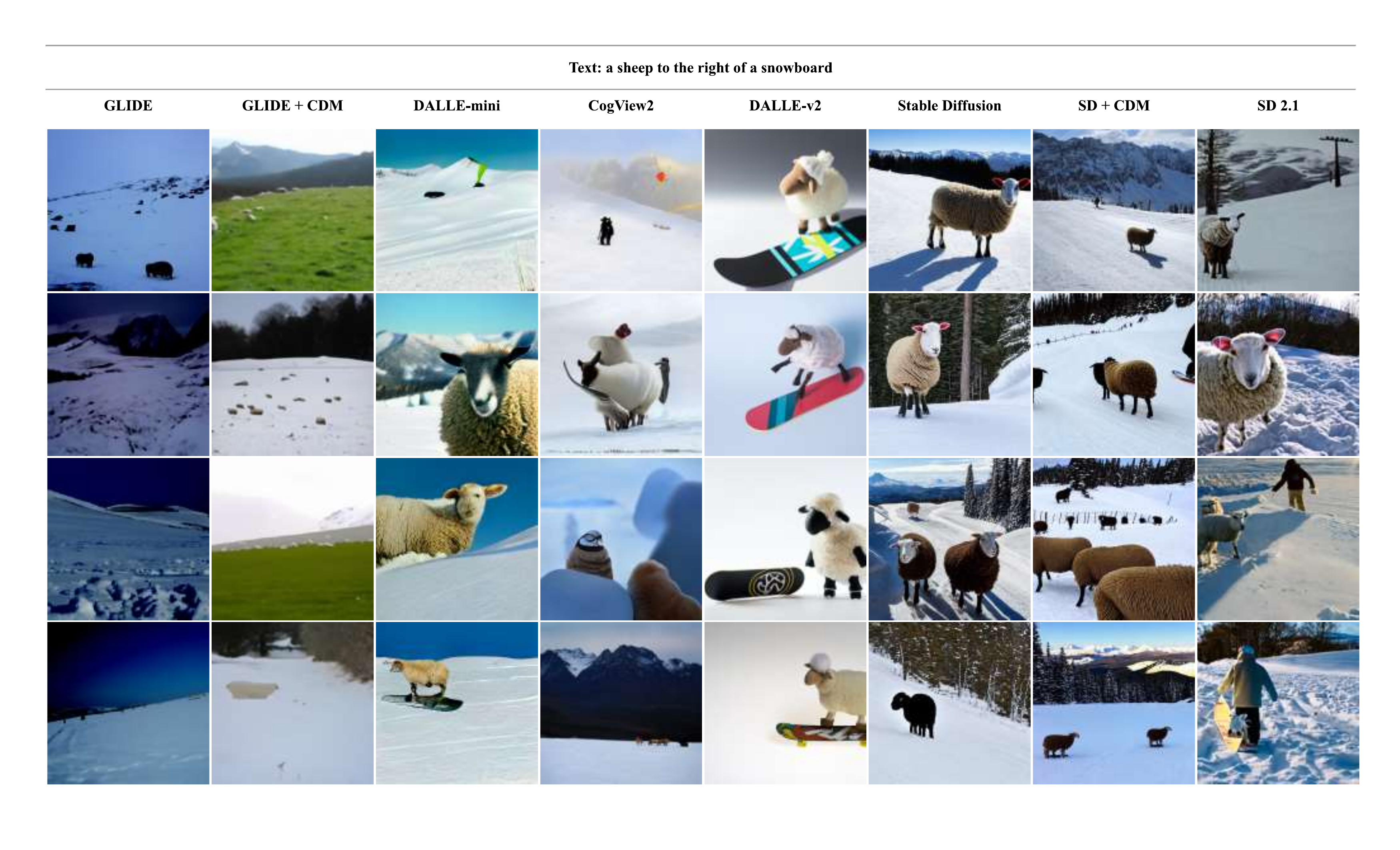}
    \caption{Illustrative examples of images generated by each of the 8 benchmark models using text prompts (top row) from the \sr\ dataset.}
    \label{fig:supp_viz_1}
\end{figure*}

\begin{figure*}
    \centering
    \includegraphics[width=0.93\linewidth]{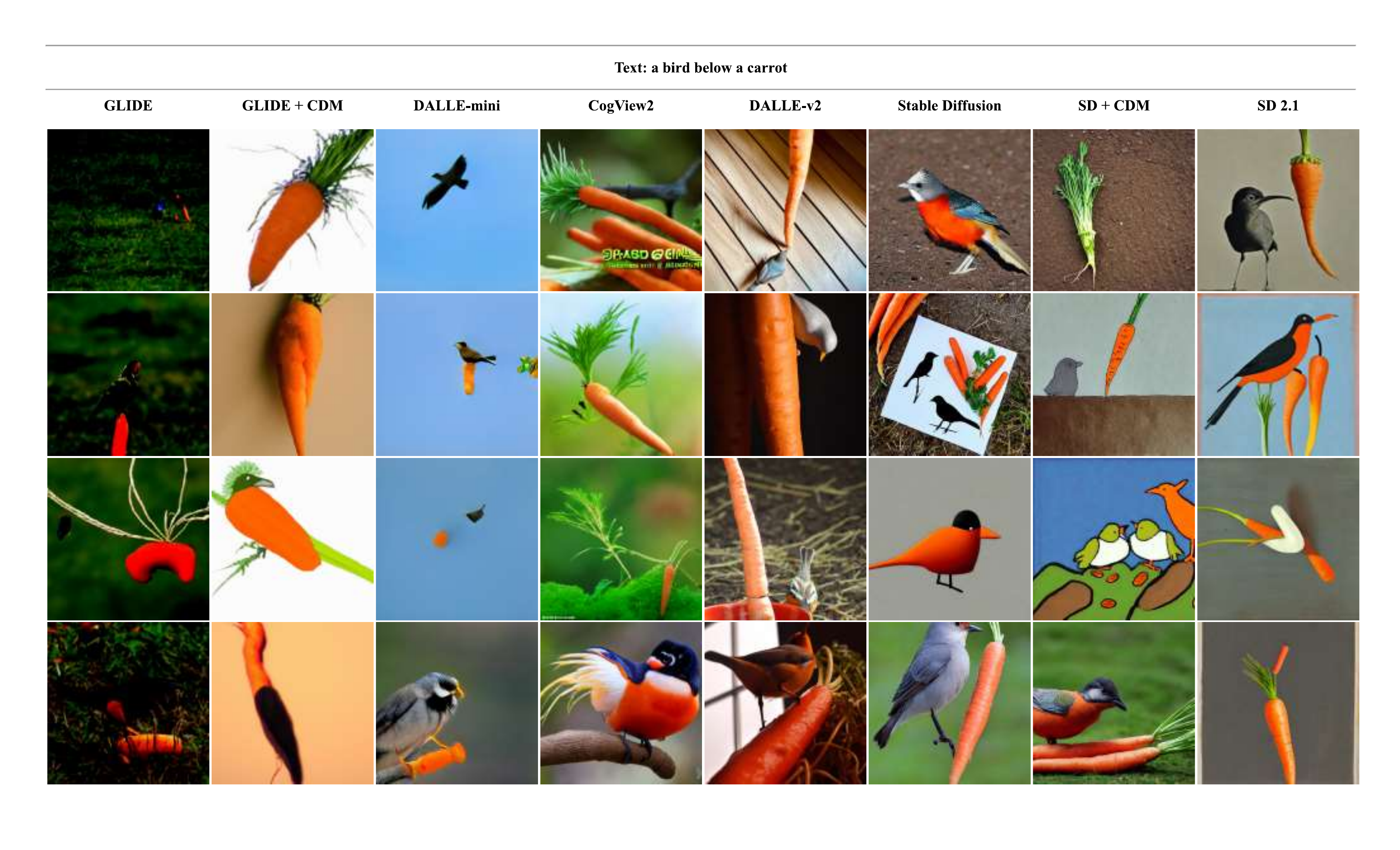}
    \caption{Illustrative examples of images generated by each of the 8 benchmark models using text prompts (top row) from the \sr\ dataset.}
    \label{fig:supp_viz_2}
\end{figure*}

\begin{figure*}
    \centering
    \includegraphics[width=0.93\linewidth]{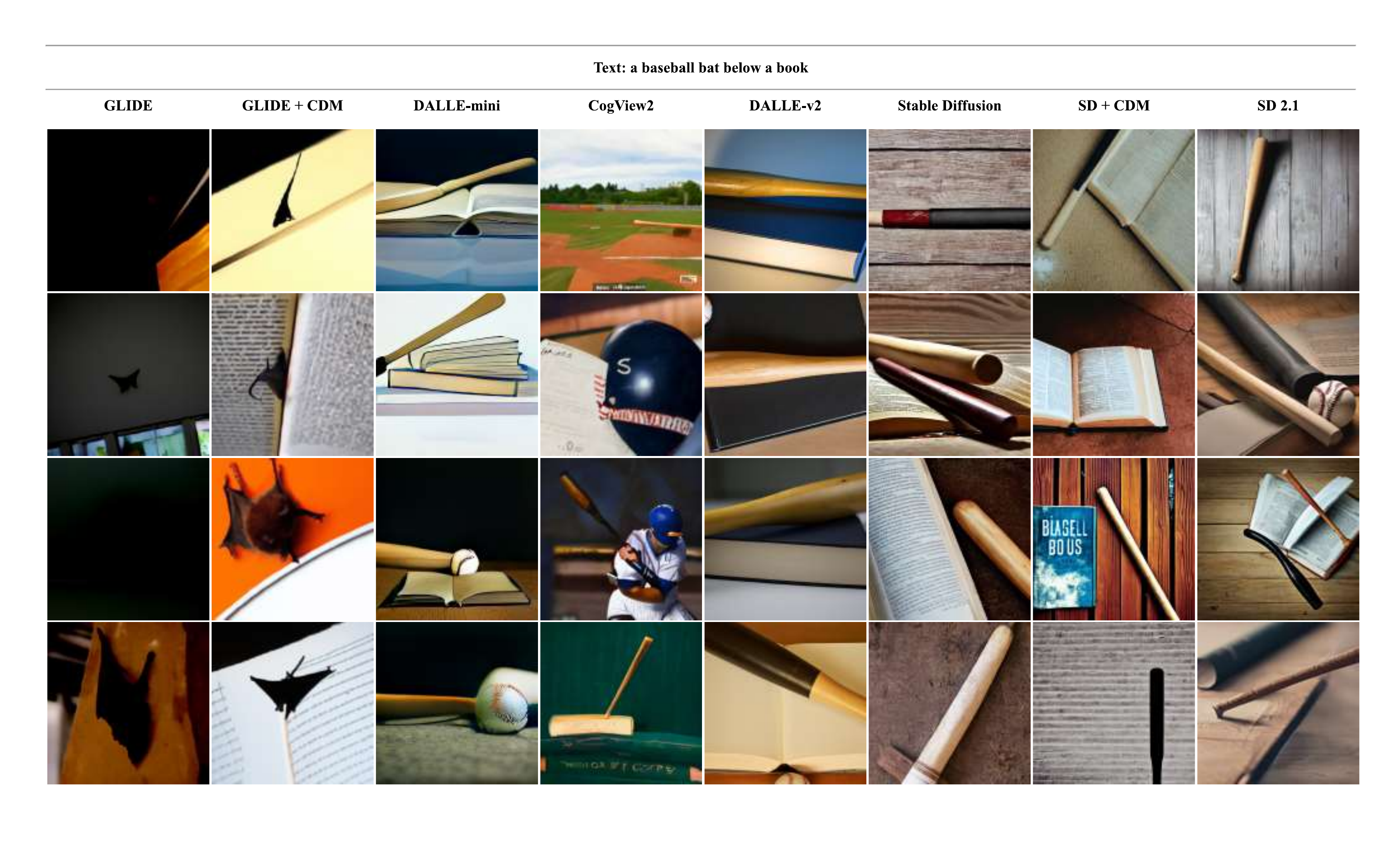}
    \caption{Illustrative examples of images generated by each of the 8 benchmark models using text prompts (top row) from the \sr\ dataset.}
    \label{fig:supp_viz_3}
\end{figure*}

\begin{figure*}
    \centering
    \includegraphics[width=0.93\linewidth]{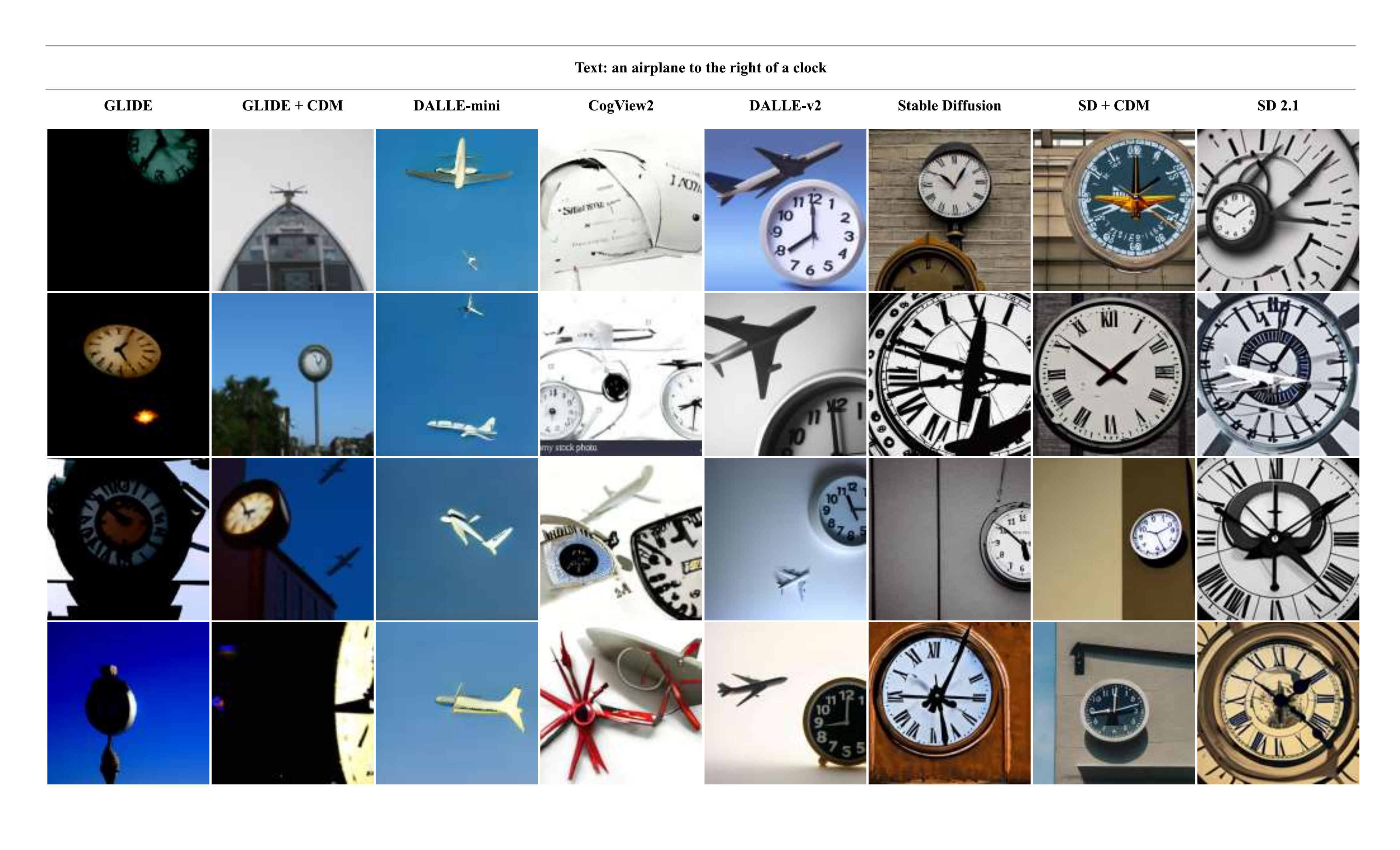}
    \caption{Illustrative examples of images generated by each of the 8 benchmark models using text prompts (top row) from the \sr\ dataset.}
    \label{fig:supp_viz_4}
\end{figure*}

\begin{figure*}
    \centering
    \includegraphics[width=0.93\linewidth]{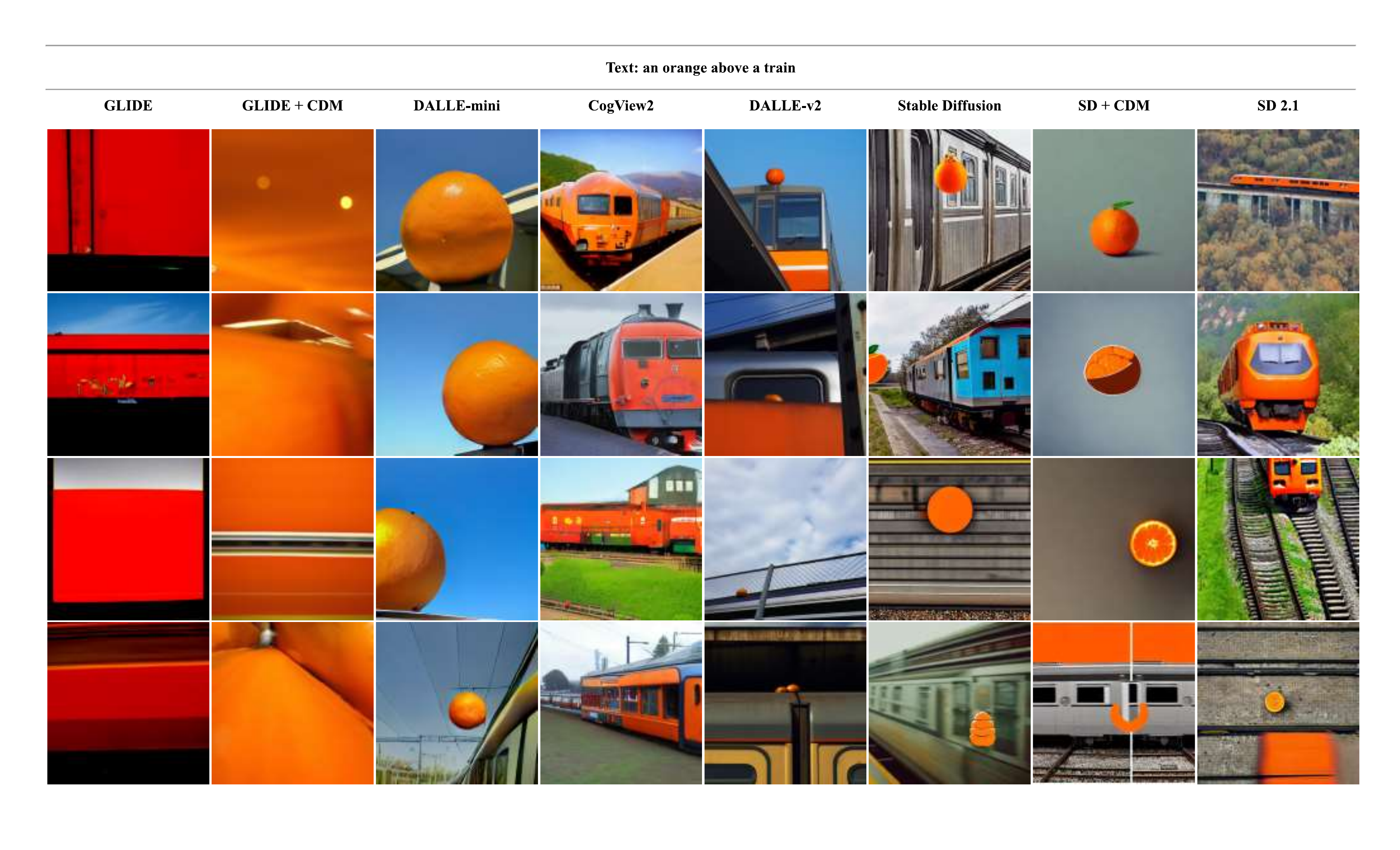}
    \caption{Illustrative examples of images generated by each of the 8 benchmark models using text prompts (top row) from the \sr\ dataset.}
    \label{fig:supp_viz_5}
\end{figure*}

\begin{figure*}
    \centering
    \includegraphics[width=0.93\linewidth]{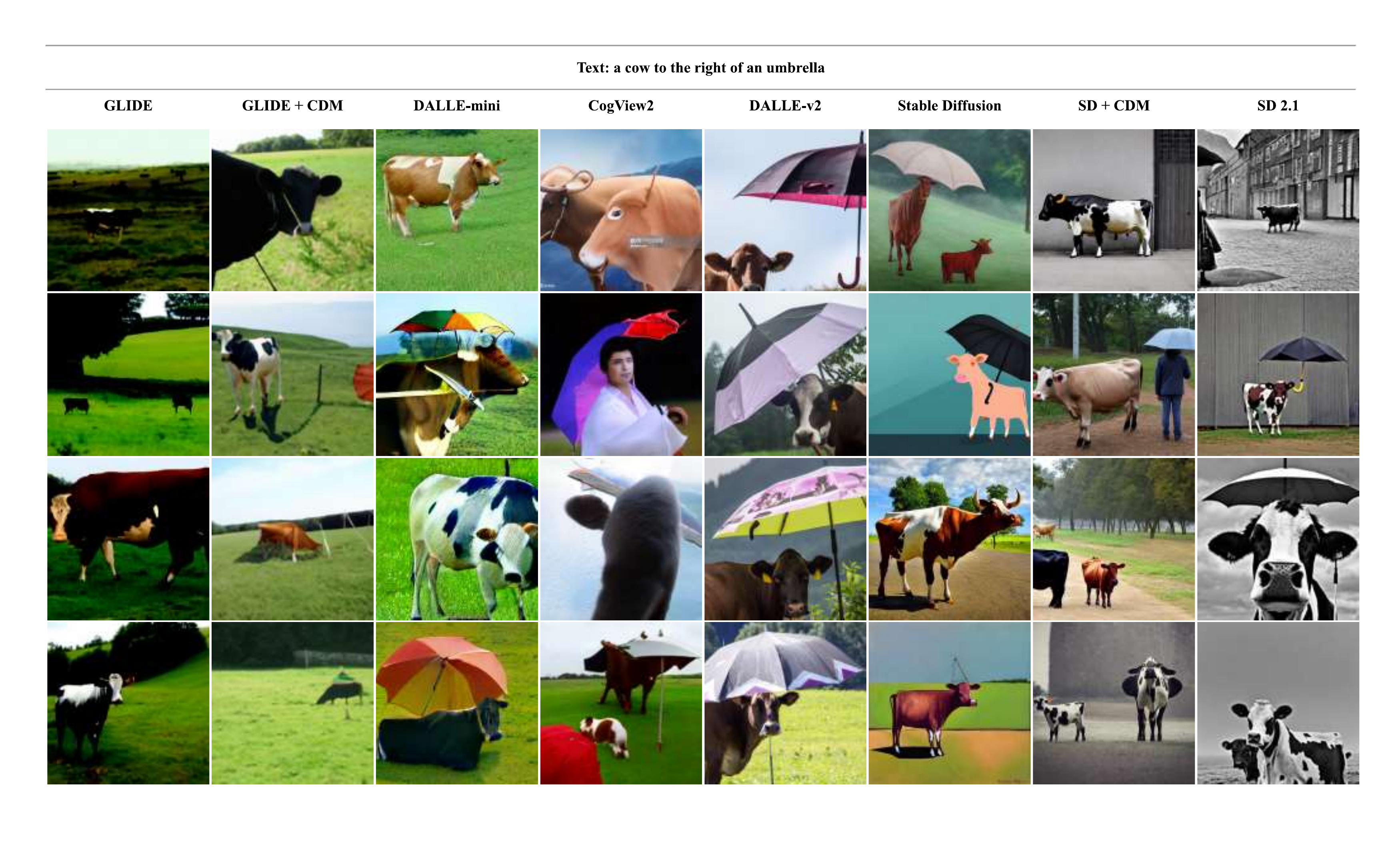}
    \caption{Illustrative examples of images generated by each of the 8 benchmark models using text prompts (top row) from the \sr\ dataset.}
    \label{fig:supp_viz_6}
\end{figure*}

\begin{figure*}
    \centering
    \includegraphics[width=0.93\linewidth]{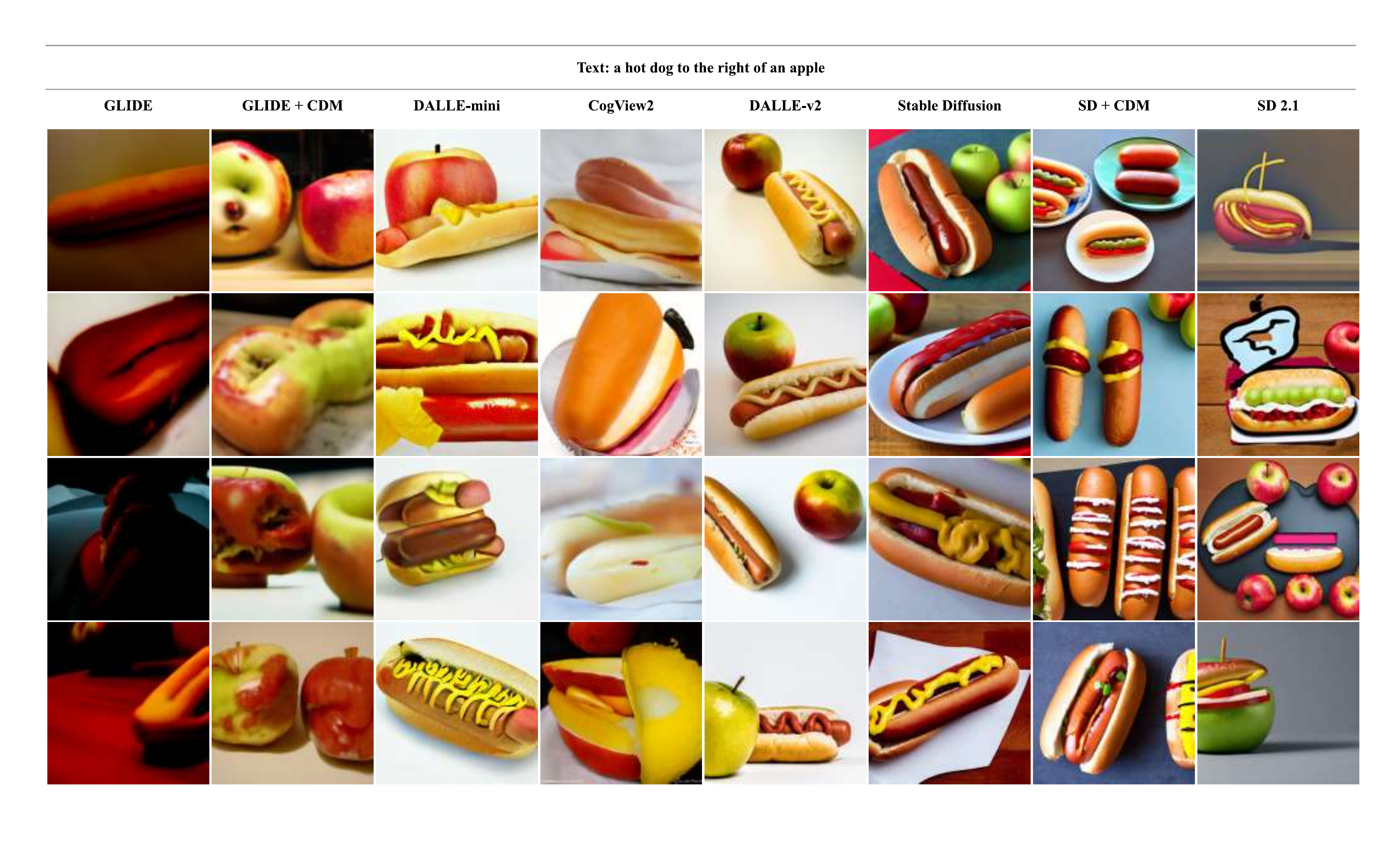}
    \caption{Illustrative examples of images generated by each of the 8 benchmark models using text prompts (top row) from the \sr\ dataset.}
    \label{fig:supp_viz_7}
\end{figure*}

\begin{figure*}
    \centering
    \includegraphics[width=0.93\linewidth]{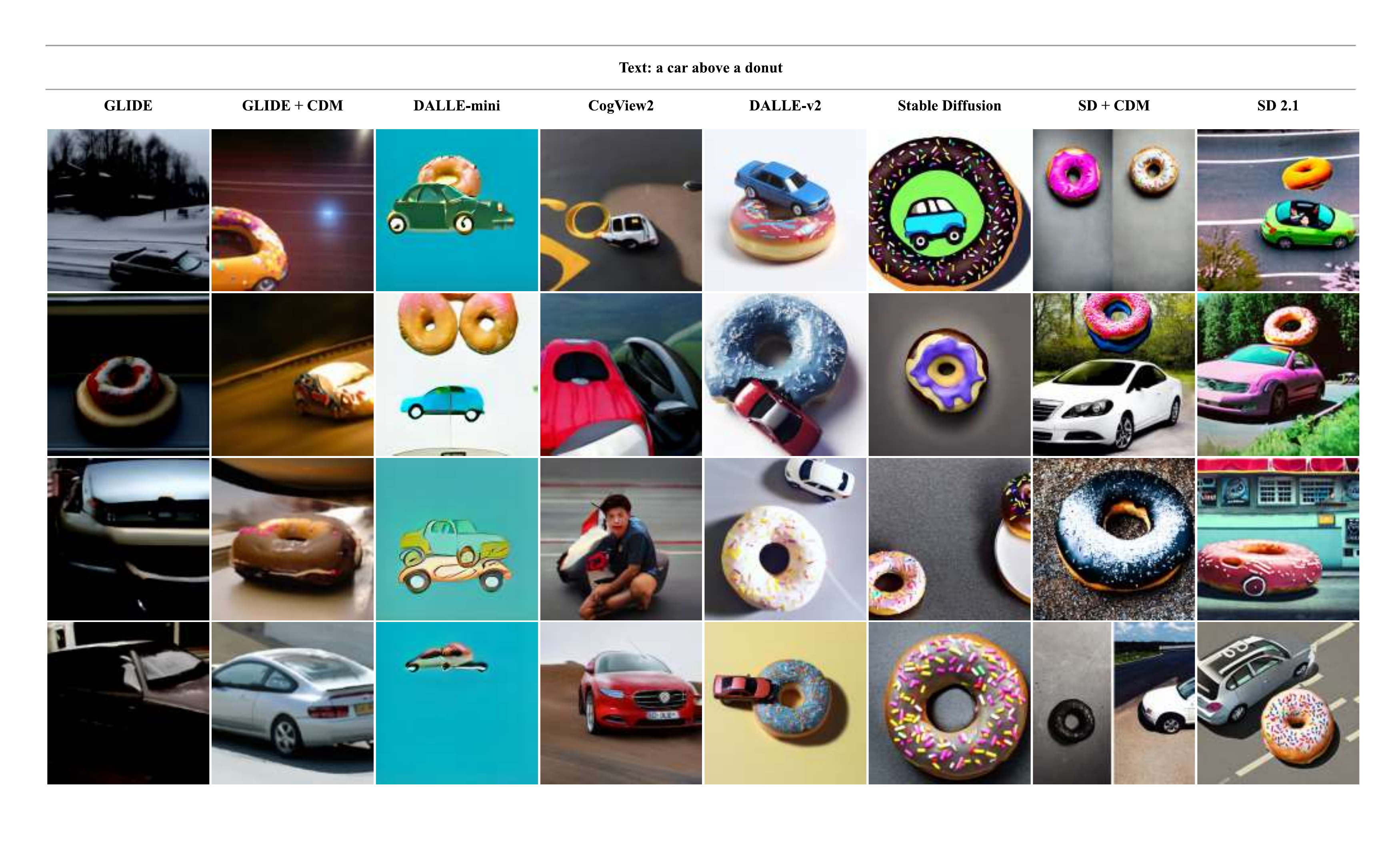}
    \caption{Illustrative examples of images generated by each of the 8 benchmark models using text prompts (top row) from the \sr\ dataset.}
    \label{fig:supp_viz_8}
\end{figure*}

We would like readers to view more examples via our anonymous project page \url{visort2i.github.io}, or the webpage \texttt{viz.html} found in the accompanying \texttt{.zip} file.
These webpages contain additional visualizations of images generated by our benchmark models for text prompts from the \sr\ dataset.
In this section we visualize some of these examples and show all $N=4$ images generated by each model, in \cref{fig:supp_viz_1,fig:supp_viz_2,fig:supp_viz_3,fig:supp_viz_4,fig:supp_viz_5,fig:supp_viz_6,fig:supp_viz_7,fig:supp_viz_8}.

\clearpage

\vfill

\end{document}